\documentclass[dvipsnames]{article}
\usepackage{colm2024_conference}
\usepackage{booktabs}
\usepackage{graphicx}
\usepackage{enumitem}
\usepackage{amsmath,amssymb,mathtools}
\usepackage{microtype}
\usepackage{caption,subcaption}
\usepackage[section]{placeins}
\usepackage{float}
\usepackage{multirow,tabularx,url,xcolor,listings,longtable,array,adjustbox,xltabular}

\usepackage{amsmath,amsfonts,bm}

\def\eqref#1{equation~\ref{#1}}

\def\1{\bm{1}}

\DeclareMathAlphabet{\mathsfit}{\encodingdefault}{\sfdefault}{m}{sl}
\SetMathAlphabet{\mathsfit}{bold}{\encodingdefault}{\sfdefault}{bx}{n}

\title{OCTAVE: A Capability-Resolved Benchmark for End-to-End HTML5 Game Generation}
\author{Kehang Li\textsuperscript{1,4,*}, Xinlan Xu\textsuperscript{3,*}, Yulin Jin\textsuperscript{2}, Heng Cao\textsuperscript{2}, Xuchen Li\textsuperscript{4},\\
Houxian Ren\textsuperscript{3}, Zihao Ji\textsuperscript{2}, Qiang Luo\textsuperscript{2}, Yadong Mao\textsuperscript{2}, Qilin He\textsuperscript{2},\\
Jian Zhang\textsuperscript{2}, Jian Zhao\textsuperscript{4}, Zihao Wang\textsuperscript{3,c}\\
\small \textsuperscript{1}Shanghai Jiao Tong University\\
\small \textsuperscript{2}Rezona Lab\\
\small \textsuperscript{3}Shandong University\\
\small \textsuperscript{4}Zhongguancun Academy\\
\textsuperscript{*}Equal contribution; \textsuperscript{c}Corresponding author}
\date{}

\newcolumntype{L}{>{\raggedright\arraybackslash}X}
\begin{document}
\maketitle
\begin{abstract}

Complete game generation requires an agent to coordinate rules, interactions, states, and presentation throughout a long-horizon development process. Reporting only a single game-level aggregate score makes it difficult to determine which game-development capabilities an agent has implemented and which types of requirements remain unmet. To address this issue, we present OCTAVE, a benchmark for complete HTML5 game generation that organizes tasks around capability coverage and diagnoses implementation differences through requirement-level evidence. We establish eight Capability Dimensions and construct 180 tasks with 1,800 Evaluation Requirements from three sources: real player requests, commercial-mechanics reconstruction, and human expert design. The benchmark covers PC Web, Mobile Web, 2D, and 3D. The requirements distinguish Completeness, which assesses publicly specified core implementation, from Richness, which assesses hidden, deeper system relationships; static source-code judgments are then used to analyze implementation performance under different task conditions. An all-task evaluation of 11 model–harness combinations yields a Completeness of 33.2\% and a Richness of 32.5\%, while 19.9\% of generation records are marked as static gate failures or missing artifacts. After excluding same-model self-evaluation, even the best-performing combination has pass rates below 50\% for both public and hidden requirements, indicating substantial gaps in both project delivery and requirement implementation. Further analysis shows that Richness is not more difficult than Completeness under every task condition. Current systems perform relatively better on Core Rules and Numerical Design, whereas AI/NPC and Feedback and Game Feel constitute more prominent capability gaps. Requirement difficulty provides a coarse-grained stratification but is only weakly negatively correlated with actual pass rates. A human cross-check on a limited sample shows that requirements judged satisfied by humans but failed by the automated evaluator constitute the primary direction of disagreement. OCTAVE thus extends the evaluation of complete game generation beyond a single aggregate score into a diagnostic evaluation localized by requirement, capability, and task condition.
\end{abstract}

\section{Introduction}

In recent years, large language models (LLMs) have evolved from code-completion systems~\citep{chen2021codex,austin2021program} into software-development agents~\citep{yang2024sweagent,wang2024openhands} capable of autonomous planning, implementation, execution, and debugging. This progress has gradually shifted the focus of automated programming research from local code generation toward the long-horizon construction of more complex software systems. Games provide a representative setting in which to study this capability. Unlike conventional programming tasks that focus primarily on standalone functions or modules, a complete game requires the coordinated implementation of core gameplay rules, state transitions, user interactions, scene organization, AI behavior, numerical systems, rendering, feedback, user interfaces, and other components. These components are tightly coupled: even when a game runs successfully and appears playable, it may still fail to implement its interaction, behavior, or presentation requirements correctly. Evaluating complete game generation therefore requires not only determining whether an agent can deliver a game artifact, but also understanding which underlying game-development capabilities it can actually implement.

Benchmarks related to game development have recently expanded from specific development tasks to complete game generation, further incorporating multimodal evaluation, behavioral verification, and runtime checks. Existing work broadly covers three directions: evaluating specific development tasks in real game engines; generating complete games from natural-language requirements and assessing their overall performance; and conducting finer-grained verification of generated-game behavior, the development lifecycle, or runtime state. These studies have advanced game-development evaluation from local tasks toward complete systems and from static inspection toward multimodal and runtime evaluation \citep{gamedevbench2026,opengame2026,gamegenverifier2026,gameasgbench2026,gamelogicbench2026}.

However, expanding the scale of evaluation does not automatically produce diagnostically useful capability coverage. A game may involve multiple capabilities simultaneously, and coverage of those capabilities may be highly uneven across tasks. Consequently, including more games or requirements does not ensure that an aggregate score can reveal an agent's specific implementation gaps. For complete game generation, the more important question is: \textbf{How can tasks be organized within an explicit capability space, and how can complete artifacts be traced back to inspectable development requirements?}

\begin{table}[H]
\caption{Comparison of existing game-development benchmarks and OCTAVE. Tasks are counted as independent game-project evaluation instances. Requirements are the specific functional requirements or scoring checklist entries accumulated across games, without repeated counting across models or runs. When only an overall evaluation is available and no game-specific functional scoring subitems are provided, each game is counted as one requirement (\textdagger{}); general quality dimensions are not counted repeatedly. GameXpert-\allowbreak Bench \cite{gamexpertbench2026} is restricted to its GameGen track. Dashes indicate that the cited papers provide insufficient information to determine the totals. GameCraft-Bench's total is counted from its released rubrics.}
\label{tab:main-1}\centering\small
\begin{tabularx}{\linewidth}{@{}>{\hsize=1.8\hsize}L >{\hsize=.9\hsize}L >{\hsize=.7\hsize}L >{\hsize=.65\hsize}L >{\hsize=.9\hsize}L >{\hsize=.9\hsize}L >{\hsize=1.15\hsize}L@{}}\toprule
Benchmark & Capability-controlled construction & Richness evaluation & Mobile Web & Complete game generation & Tasks & Requirements \\
\midrule
\href{https://arxiv.org/html/2602.11103v2\#S2}{GameDevBench \cite{gamedevbench2026}} & No & No & No & No & 333 & \textemdash{} \\
\href{https://arxiv.org/html/2604.18394v2\#S4}{OpenGame-Bench \cite{opengame2026}} & No & No & No & Yes & 150 & \textemdash{} \\
\href{https://arxiv.org/html/2509.20136v1\#S2.SS2}{V-GameGym \cite{vgamegym2025}} & No & No & No & Yes & 2,219 & 2,219\textdagger{} \\
\href{https://arxiv.org/html/2605.07442v1\#S5}{GameGen-Verifier \cite{gamegenverifier2026}} & No & No & No & Yes & 100 & \textemdash{} \\
\href{https://arxiv.org/html/2606.17861v1\#S4}{GameCraft-Bench \cite{gamecraftbench2026}} & No & Yes & No & Yes & 140 & 2,305 \\
\href{https://arxiv.org/html/2605.17637v2\#S3.SS2}{WebGameBench \cite{webgamebench2026}} & No & No & No & Yes & 111 & 111\textdagger{} \\
\href{https://arxiv.org/html/2608.21833\#S4.SS1}{GameXpert-\allowbreak Bench \cite{gamexpertbench2026}} & No & Yes & No & Yes & 97 (GameGen) & \textemdash{} \\
\href{https://arxiv.org/html/2607.03525v2\#S1}{GameEngine-\allowbreak Bench \cite{gameenginebench2026}} & No & No & No & No & 110 & \textemdash{} \\
OCTAVE & Yes & Yes & Yes & Yes & 180 & 1,800 \\
\bottomrule\end{tabularx}\end{table}

OCTAVE organizes benchmark construction around this question. Drawing on the game-design concepts in \emph{Elements of Game Design}~\citep{zubek2020elements}, we operationalize eight Capability Dimensions (C1–C8) spanning rules, AI, numerical design, interaction, scenes, rendering, feedback, and guidance. We then construct tasks from three sources: real player requests, commercial-mechanics reconstruction, and human expert design. From the construction stage onward, tasks are separated into PC Web and Mobile Web tracks and cover both 2D and 3D. Each task is further associated with Completeness Requirements for publicly specified core implementation and Richness Requirements for hidden, deeper system relationships, connecting capability coverage to specific source-code requirements.

The experimental results show substantial gaps in current systems with respect to both project delivery and requirement implementation; after excluding same-model self-evaluation, no combination achieves a pass rate above 50\% for either public or hidden requirements. Completeness and Richness have similar overall pass rates, while their point estimates vary by task source and track under the corresponding evaluation protocols. Current systems perform relatively better on C1 Core Rules and C3 Numerical Design, whereas AI/NPC and Feedback and Game Feel have lower combined pass rates. Requirement-level and multilayer decompositions therefore provide more than an overall leaderboard: they distinguish among delivery reliability, completion of explicit functionality, and deeper system relationships. In the human cross-check on a limited sample, the principal disagreement consists of requirements that humans judge satisfied but the automated evaluator fails, revealing systematic differences between static judgment and human observation.

OCTAVE complements runtime verification and player-experience evaluation for complete game generation. This study uses static source-code evidence to identify gaps in requirement implementation, focusing on how capability coverage connects to inspectable task requirements and how task source, task stratification, and gate outcomes affect the interpretation of results. Project code and released data are available in the \href{https://github.com/rezona-ai/octave-bench}{OCTAVE public repository}. Released materials and dataset documentation are listed in Appendix H. Our main contributions are as follows:

\begin{itemize}

\item \textbf{Capability-first task organization.} We use the Capability Dimensions to guide task coverage and distinguish task-level multilabel capability annotation from the primary capability assignment of a requirement.

\item \textbf{A multisource benchmark for complete HTML5 games.} We construct 180 tasks and 1,800 requirements covering three sources, two platforms, 2D/3D, four requirement-difficulty levels, publicly specified core requirements, and hidden, deeper requirements.

\item \textbf{Requirement-level diagnosis.} For 11 model–harness combinations, we report results by source, capability, difficulty, platform, and gate support set, and use a human cross-check on a limited sample to bound the interpretation of static judgments.

\end{itemize}

\begin{figure}[H]\centering\includegraphics[width=0.95\linewidth]{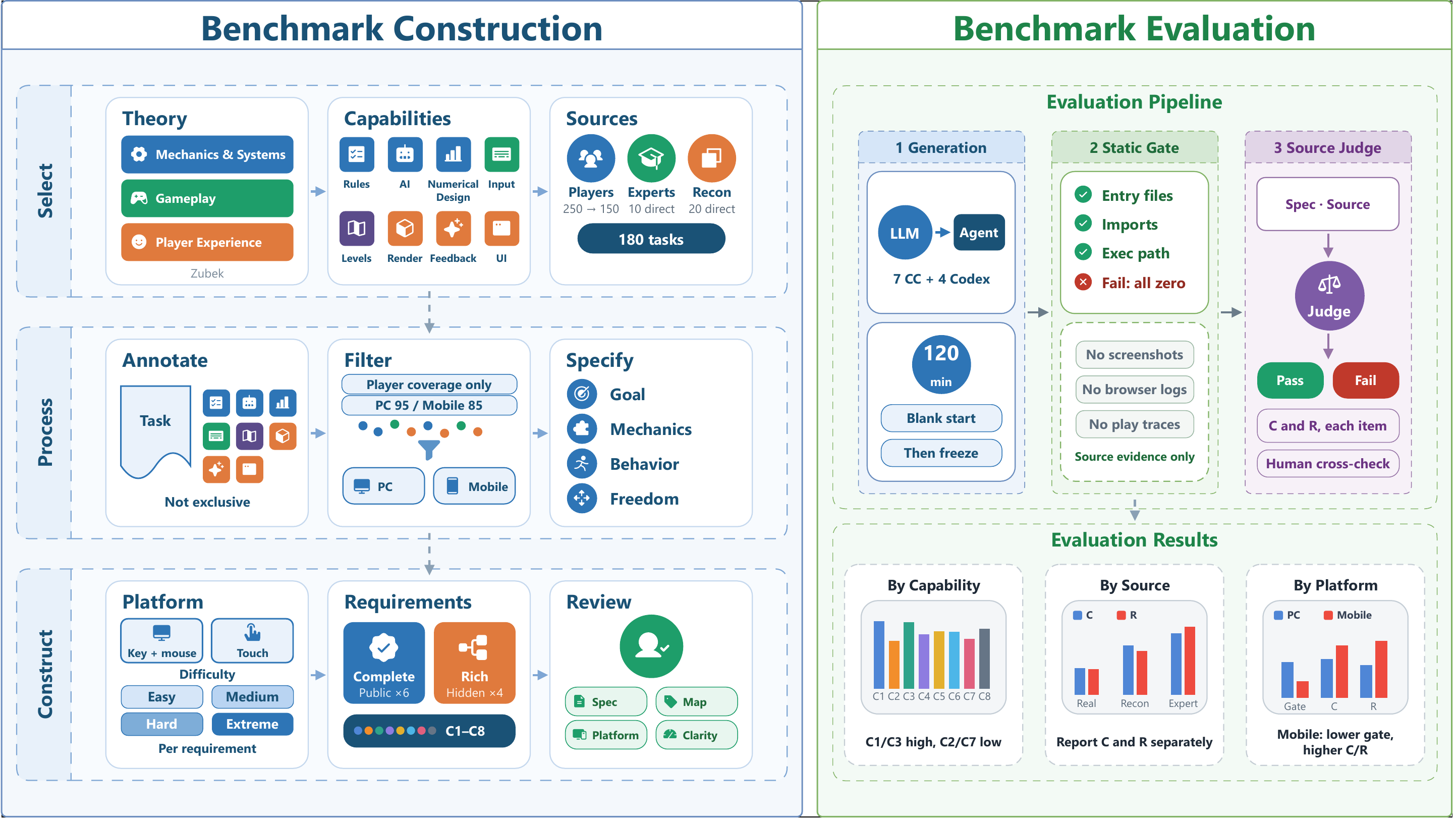}\caption{Construction and evaluation pipeline}\label{fig:F00_pipeline}\end{figure}

\section{Related Work}

\subsection{Evaluation of LLM Programming Capabilities}

LLM programming benchmarks have gradually expanded from local code generation to long-horizon software development. Function-level benchmarks such as HumanEval \cite{chen2021codex} and MBPP \cite{austin2021program} primarily evaluate standalone programming problems by executing tests, abstracting project-level dependencies away from the task. At the repository level, RepoBench \cite{liu2023repobench} evaluates cross-file code completion in real code repositories, whereas SWE-bench \cite{jimenez2024swebench} evaluates agents' ability to modify and repair existing projects using real GitHub Issues and repository tests, a setting addressed by agent systems such as SWE-agent \cite{yang2024sweagent} and OpenHands \cite{wang2024openhands}. More recently, research has begun to examine the construction of complete software from scratch. Commit0 \cite{zhao2024commit0} evaluates library implementation from natural-language specifications; PaperBench \cite{starace2025paperbench} asks agents to replicate papers' research contributions by implementing codebases and executing experiments; and NL2Repo-Bench \cite{nl2repo2026} goes further by requiring complete software repositories to be constructed from a single natural-language document without project scaffolding. These settings progressively increase the scope of system integration required by a task, from standalone functions to complete software systems. Game development is among the more challenging such settings because a game simultaneously involves gameplay logic, state transitions, user interaction, rendering, scene organization, and mutually coupled runtime systems.

\subsection{Game Generation and AI-Assisted Development}

Research on Procedural Content Generation (PCG) \cite{hendrikx2013pcg} has long used automated methods to generate content at different levels of games, including objects, spaces, systems, scenes, and designs. PCGML \cite{summerville2018pcgml} uses machine-learning models trained on existing content to generate functional game content such as levels, maps, and stories. With the development of LLMs, research has gradually shifted from generating individual content elements toward AI-assisted game development and end-to-end game construction. GameGPT \cite{chen2023gamegpt} assigns planning, task decomposition, and implementation to multiple specialized agents; Instruction-Driven Game Engines (IDGE) \cite{wu2024idge} models gameplay as next-state prediction conditioned on natural-language rules, previous states, and player actions, demonstrated primarily on poker variants. RPGAgent \cite{rpgagent2026} further turns story outlines into playable game prototypes through agents specializing in narrative, spatial layout, and gameplay mechanics. OpenGame \cite{opengame2026} targets end-to-end Web game generation, constructing complete games directly from high-level designs and using template reuse and debugging capabilities to address cross-file consistency, scene connections, runtime integration, and related problems. These studies reflect the broader movement of game AI from content generation toward complete game construction, while also establishing games as an important setting for evaluating long-horizon code generation and agentic software-development capabilities.

\subsection{Game-Development Benchmarks}

In parallel with these generation systems, game-development benchmarks have continued to broaden the scope of evaluation. GameDevBench \cite{gamedevbench2026} and GameEngine-\allowbreak Bench \cite{gameenginebench2026} evaluate specific development tasks in real game-engine environments, covering functionality such as gameplay, graphics, UI, AI, and animation. V-GameGym \cite{vgamegym2025}, OpenGame-Bench \cite{opengame2026}, GameCraft-Bench \cite{gamecraftbench2026}, and WebGameBench \cite{webgamebench2026} move further toward complete game generation, evaluating generated results through multimodal assessment, interaction replay, browser execution, and other methods. Recent work has also introduced finer-grained evaluations of behavior and the development lifecycle. GameGen-Verifier \cite{gamegenverifier2026} decomposes a Game Specification into independently verifiable behaviors; GameXpert-Bench \cite{gamexpertbench2026} covers development stages including generation, repair, and optimization; and GameASG-Bench \cite{gameasgbench2026} and GameLogicBench \cite{gamelogicbench2026} further introduce explicit runtime-behavior checks. Some studies also use functional dimensions to characterize tasks. For example, WebGameBench \cite{webgamebench2026} decomposes browser-game functional requirements along behavior, spatial, and temporal/state axes. GameCraft-Bench \cite{gamecraftbench2026} and GameXpert-Bench \cite{gamexpertbench2026} assess implementation beyond core functionality through content-depth scoring and bonus-event evaluation, respectively. Overall, existing benchmarks still exhibit clear coverage gaps. Most existing benchmarks do not use a unified system of game-development capabilities to guide task construction or systematically evaluate hidden, deeper system relationships (Richness), while dedicated Mobile Web evaluation remains uncommon. Moreover, some benchmarks focus on isolated development tasks or game-level overall evaluation, making it difficult to trace results to specific functional requirements and their capability dimensions under a unified protocol. OCTAVE organizes the C1–C8 generation capabilities from a game-design perspective and uses them for task construction, coverage, and requirement design. Static source-code diagnosis and runtime-behavior verification are complementary.

\section{Benchmark Construction}

OCTAVE is designed to systematically evaluate the ability of large models to generate complete HTML5 games from a Game Specification, defined as the natural-language game requirements provided to the agent. We seek to answer two related questions: \textbf{Which game-development capabilities must a model implement, and to what extent are these capabilities implemented across different game requirements?} We therefore use structured Capability Dimensions as the organizing principle of construction. We first establish a capability framework grounded in game-design theory, collect candidate tasks from complementary sources, and select the final tasks according to capability coverage. We then normalize the tasks into Game Specifications, construct Evaluation Requirements corresponding to the capability dimensions, and finally control benchmark quality through review by project experts. Through this process, the capability framework is used not only to analyze model-generated results but also to constrain benchmark construction beginning at the task-selection stage.

\subsection{Capability Dimensions}

To establish a structured capability space for complete game generation, we draw on the game-design concepts proposed in \emph{Elements of Game Design}~\citep{zubek2020elements}. This framework discusses game design across interrelated layers such as \textbf{Mechanics and Systems, Gameplay, and Player Experience}, and further addresses specific design concepts including Mechanics, Systems, Gameplay, and Macrostructure. Based on the implementation objectives of complete game-generation tasks, we operationalize these concepts as development capabilities that can be evaluated through game requirements and observable behavior, rather than directly adopting the theoretical concepts as evaluation categories.

Specifically, Mechanics corresponds to game rules, state transitions, and core mechanics, yielding \textbf{C1 Core Rules}. Systems concerns autonomous behavior and numerical relationships within a game and is further divided into \textbf{C2 AI/NPC} and \textbf{C3 Numerical Design}. Gameplay concerns the actual interaction between the player and the game system and corresponds to \textbf{C4 Interaction and Input}. Macrostructure concerns level organization, scene structure, and game progression and corresponds to \textbf{C5 Levels and Scenes}. For those aspects of Player Experience that can be described and judged through implementation requirements, we further establish \textbf{C6 Animation and Rendering, C7 Feedback and Game Feel, and C8 Guidance and Meta-UI}, covering visual presentation, event feedback, and player-facing interfaces and guidance, respectively. The notion of \emph{game feel} follows its standard use for moment-to-moment control feedback and polish~\citep{swink2008gamefeel}. In this way, we translate game-design concepts into eight Capability Dimensions for complete game generation.

\begin{table}[H]
\caption{Capability Dimensions and their relationship to game-design concepts.}
\label{tab:capabilities}\centering\small
\begin{tabularx}{\linewidth}{*{3}{L}}\toprule
Reference design concept & Capability Dimension & Operational scope \\
\midrule
Mechanics & C1 Core Rules & Game rules, state transitions, core mechanics, and win/loss conditions \\
Systems & C2 AI/NPC & NPC behavior, decision logic, and autonomous system behavior \\
Systems & C3 Numerical Design & Attributes, resources, economies, progression, and numerical relationships \\
Gameplay & C4 Interaction and Input & Player input, interaction mapping, and game controls \\
Macrostructure & C5 Levels and Scenes & Spatial organization, levels, scenes, and game progression \\
Player Experience & C6 Animation and Rendering & Rendering, animation, and dynamic visual states \\
Player Experience & C7 Feedback and Game Feel & Visual, auditory, and state feedback related to actions and game events \\
Player Experience & C8 Guidance and Meta-UI & Tutorials, prompts, HUDs, menus, pausing, and end states \\
\bottomrule\end{tabularx}\end{table}

This framework further decomposes the holistic question of whether a game was generated into independently examinable development capabilities spanning rules and systems, player interaction, game structure, and player-facing presentation, thereby providing a unified basis for subsequent capability-first task construction.

\subsection{Task Collection}

Guided by this capability framework, we construct candidate tasks from three complementary sources to capture the distinct perspectives on capability coverage provided by real player requests, public game mechanics, and expert design.

The first source is \textbf{real player requests}. Starting from game concepts and gameplay requirements proposed by players, we transform natural-language requests into candidate tasks with explicit goals, system relationships, and judgeable behaviors through cleaning, normalization, and mechanical elaboration. This process preserves the themes and gameplay characteristics of the original requests while making them suitable for analyzing and evaluating different game-development capabilities. Request mining, candidate formation, and source processing are described in Appendix A.1.

The second source is \textbf{public game ecosystems such as Steam and mobile-game stores}. We extract representative core mechanics from publicly documented games and reorganize them into independent game requirements, rather than asking models to reproduce the original games in full. Compared with requests supplied directly by players, these tasks introduce game designs with explicit mechanical structures and system relationships, providing the capability framework with coverage from a different source. Information sources, mechanics reorganization, and selection methods are described in Appendix A.2.

The third source is \textbf{human expert design}. Experts construct tasks based on their experience in game design and development to supplement complex system relationships and engineering requirements that are not adequately represented in real requests or public game mechanics. From the design stage onward, these tasks follow the unified capability framework and evaluation protocol and are included in the final task set. The specific design principles are described in Appendix A.3.

These three sources expand the range of task content and system relationships, allowing the dataset to represent diverse design requirements.

\subsection{Capability-Driven Selection and Specification}

Collecting tasks from multiple sources expands the range of candidate content but does not guarantee sufficient and balanced coverage of each game-development capability. During task selection, we therefore first annotate candidate tasks by capability and then select tasks derived from real player requests primarily according to the balance of their capability coverage. Commercial-mechanics reconstruction and expert-designed tasks are incorporated directly into the final set according to their respective construction protocols. Because a game task generally involves multiple interrelated development capabilities, tasks are described with multiple labels to preserve the combinations of capabilities they actually involve. The statistical protocol for task labels and capability coverage before and after selection are provided in Appendix A.4.

In addition to capability coverage, we consider the development constraints imposed by the target platform. \textbf{PC Web and Mobile Web are separated into independent tracks from the task-construction stage onward}. Both target browser-based HTML5 games, but PC Web is designed for desktop keyboard-and-mouse input and viewport conditions, whereas Mobile Web is designed for touch input, small-screen layouts, and environments without a keyboard. This platform distinction enables a more comprehensive examination of large models' game-production capabilities across environments. Platform composition and the track-separation procedure are described in Appendix A.5.

After task selection, we normalize candidate requirements into a unified \textbf{Game Specification}. Each specification describes the game's objectives, core mechanics, required behaviors, and design constraints from the perspectives of the player and game design. This gives tasks from different sources a consistent form of expression while preserving reasonable implementation freedom. On the one hand, requirements must be explicit enough to support behavior-level judgments. On the other hand, the specification does not prescribe a fixed code structure or implementation template, thereby avoiding an evaluation limited to adaptation to a particular form of engineering organization. Detailed construction guidelines and examples are provided in Appendix B.1–B.4.

\subsection{Evaluation Requirements and Rubric Construction}

The Capability Dimensions define the game-development capabilities that the benchmark needs to cover, but capability categories alone cannot directly determine whether a model has actually implemented those capabilities. To make the capability framework operational for evaluation, we derive \textbf{Evaluation Requirements} from each Game Specification and use specific, observable behavioral requirements as the units of judgment. For example, in a task containing destructible walls and enemy pathfinding, a Completeness Requirement may check whether a newly opened route becomes usable after a wall is destroyed, whereas a Richness Requirement may further check whether paths and states remain consistent when multiple walls are destroyed in succession or when an enemy is moving along an outdated route. This example illustrates how the two requirement types respectively target the core functional loop and system relationships inferred from the task description. Because a complete game must both satisfy explicitly stated functional requirements and maintain reasonable behavior under coordination among multiple systems and complex states, we divide Evaluation Requirements into two complementary categories: Completeness Requirements and Richness Requirements.

\textbf{Completeness Requirements} correspond directly to mandatory content in the Game Specification and are used to determine whether the model has completed a task's core functionality. Each task has a set of Completeness Requirements described in terms of observable behavior or outcomes. For example, an Evaluation Requirement does not merely state that a system "should exist"; it further specifies which observable outcome that system should produce under particular conditions. In this way, the evaluation focuses on whether functionality actually forms a complete behavioral loop, rather than merely whether related implementation appears in the code.

\textbf{Richness Requirements} are designed to assess whether a model can implement deeper system relationships that can be naturally inferred from the specification. They do not add new basic requirements that the model must complete, but instead examine cross-system coupling, edge cases, complex states, and stable behavior under changes in scale. Richness Requirements follow a zero-prompting design: they are omitted from the generation prompt. They therefore reflect the model's ability to go beyond explicit requirements and construct a robust, complete game system, rather than merely its instruction-following ability.

To support capability analysis using requirement-level judgments, we associate Evaluation Requirements with C1–C8 so that scores can be aggregated by Capability Dimension. A single requirement may be associated with multiple capabilities; in the main analysis, we use its first capability label as an aggregation convention for primary assignment. The observed label ordering and alternative aggregation protocols are documented in Appendix G.2. Requirement-construction guidelines are provided in Appendix B.1–B.5.

In addition to capability coverage, requirement difficulty is an important dimension for understanding evaluation results. During construction, we assess design complexity based on each requirement's intrinsic attributes, considering the breadth of system coupling, algorithmic and simulation core, verification rigor, dynamic resource governance and adaptation, multi-agent and social cognition, and scale and performance. The score definition, scoring anchors, and requirement-stratification rules are detailed in Appendix C, and the overall distribution is shown in Section 3.6.

\subsection{Human Validation}

The Capability Dimensions provide a unified basis for task selection and Evaluation Requirements, but their effective application still depends on the clarity of the specifications, the reasonableness of capability mappings, and the judgeability of the requirements. Before formal evaluation, a group of project experts with backgrounds in game development, design, and testing therefore reviewed the tasks and their Evaluation Requirements. The review focused on whether the requirements were clear and feasible, whether requirements were appropriately mapped to capability dimensions, whether platform interaction paths were complete, and whether task complexity matched the design objectives.

This review controlled the quality of benchmark construction and reexamined ambiguities, omissions, and inappropriate designs in tasks and requirements. The reviewers' backgrounds, review scope, and related records are provided in Appendix B.6–B.7.

We subsequently revised the dataset based on the review results, for example by adding touch-input paths for Mobile Web, clarifying edge conditions, and improving judgeability. Detailed revision statistics are provided in Appendix B.5 and Appendix B.7.

\subsection{Dataset Statistics}

After multisource task collection, capability-driven selection, specification normalization, and expert review, OCTAVE yields a task set for complete HTML5 game generation. This section summarizes the final benchmark's scale, sources, game genres, visual forms, platforms, and requirement difficulty, and further presents its coverage of the Capability Dimensions.

\begin{table}[H]
\caption{Overall statistics of OCTAVE. Candidate tasks are the 280 tasks entering final selection: 250 player requests, the 20 retained commercial-mechanics reconstructions, and the 10 expert-designed tasks. A further 11 reconstruction drafts were discarded upstream and are not counted.}
\label{tab:main-2}\centering\small
\begin{tabularx}{\linewidth}{@{}>{\hsize=.75\hsize}L >{\hsize=1.25\hsize}L@{}}\toprule
Statistic & Value \\
\midrule
Candidate tasks / final tasks & 280 / 180 \\
Task sources & 150 player requests; 20 commercial-mechanics reconstructions; 10 expert-designed tasks \\
Game genres & 6 \\
Visual forms & 2D, 3D \\
Target platforms & 95 PC Web; 85 Mobile Web \\
Requirements per task & 6 Completeness; 4 Richness \\
Total requirements & 1,800 \\
Completeness Requirements / Richness Requirements & 1,080 / 720 \\
Capability Dimensions & C1–C8 \\
Requirement-difficulty stratum & Easy, Medium, Hard, Extreme \\
\bottomrule\end{tabularx}\end{table}

\begin{table}[H]
\caption{Category distribution of the final tasks. Each classification dimension is counted separately, and each dimension sums to 180 tasks.}
\label{tab:main-3}\centering\small
\begin{tabularx}{\linewidth}{*{3}{L}}\toprule
Dimension & Category & Tasks \\
\midrule
Source & Player request & 150 \\
 & Commercial-mechanics reconstruction & 20 \\
 & Expert design & 10 \\
Game genre & Action & 30 \\
 & Adventure \& Puzzle & 31 \\
 & RPG & 24 \\
 & Simulation \& Sandbox & 45 \\
 & Social \& Tabletop & 16 \\
 & Strategy & 34 \\
Visual form & 2D & 91 \\
 & 3D & 89 \\
Platform & PC Web & 95 \\
 & Mobile Web & 85 \\
\bottomrule\end{tabularx}\end{table}

\begin{figure}[H]\centering\includegraphics[width=0.95\linewidth]{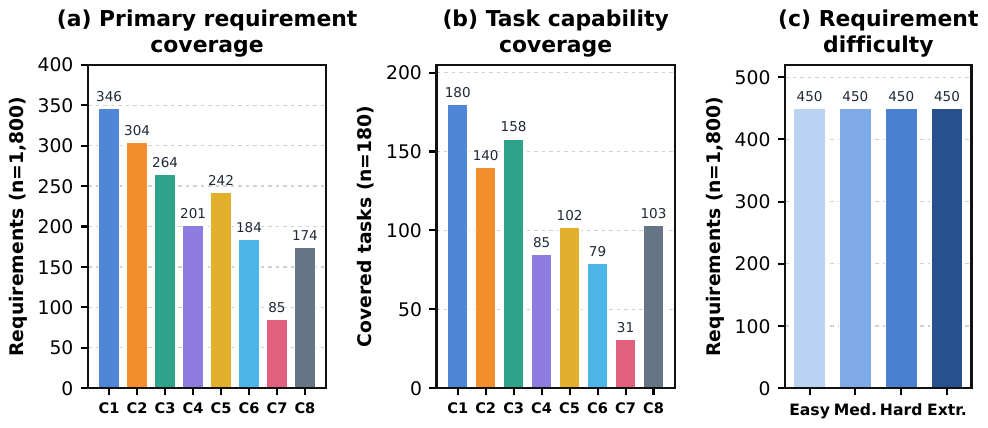}\caption{(a) Distribution of primary requirement capabilities; (b) task-level multilabel capability coverage; and (c) distribution of requirement difficulty. The primary-capability distribution uses the first capability label of each Evaluation Requirement; task-level multilabel coverage reflects the occurrence of different capabilities in the task set; and the requirement-difficulty distribution shows the number of requirements in the four equal-count strata defined in Appendix C. Detailed label summaries are provided in Appendix A.4, and the difficulty-stratification rules are provided in Appendix C.}\label{fig:F01_dataset_formal_cn}\end{figure}

\section{Experiments}

\subsection{Experimental Setup}

We evaluate 11 model–harness combinations that use Claude Code~\citep{claudecode2025} or Codex CLI~\citep{codexcli2025} to generate complete HTML5 games from natural-language Game Specifications. The models are DeepSeek-V4.1-Flash~\citep{deepseekv41flash2026}, DeepSeek-V4-Flash-0731 and DeepSeek-V4-Pro~\citep{deepseekv42026}\footnote{The 0731 suffix denotes the 31 July 2026 snapshot. The cited technical report describes the DeepSeek-V4-Flash and DeepSeek-V4-Pro series.}, Fable~5~\citep{fable52026}, Fable~5.1~\citep{fable512026}, Opus~5~\citep{opus52026}, GLM-5.2~\citep{glm522026}, GLM-5.3~\citep{glm532026}, GPT-5.5~\citep{gpt552026}, GPT-5.6-sol~\citep{gpt562026}, and GPT-6-astra~\citep{gpt6astra2026}. All combinations use the same task set and the corresponding generation protocol within each track. Evaluation-call differences across tracks are described in Appendix E.1. Because different combinations use different models and agent harnesses, we report performance at the combination level. Reasoning-effort settings, runtime configurations, and budgets are provided in Appendices D.1–D.3.

The experiment consists of generation and evaluation stages. The agent receives only the final Game Specification and develops autonomously from an empty workspace. After generation, the evaluation system inspects the final artifact and assesses its implementation against the preconstructed Evaluation Requirements.

\subsection{Generation Protocol}

During generation, each task begins in an independent workspace containing no preexisting game logic or task-specific code. Given a single initialization instruction, the agent autonomously completes development, and the final workspace is retained as the evaluation artifact. Budgets, permissions, timing, retry rules, and the generation protocol are described in Appendices D.2–D.3.

\subsection{Automated Evaluation}

At the evaluation stage, we perform a two-stage static source-code review. The first stage determines whether the generated project can run, and the second evaluates each game function and system behavior against its requirement. Both stages use gpt-6-astra~\citep{gpt6astra2026} as the LLM evaluator~\citep{zheng2023llmjudge}.

\textbf{Stage 1: Static Runnability Gate.} The evaluator examines the submission directory, required files, entry artifacts, and relevant execution paths to determine whether the generated project can run and whether it supports subsequent evaluation.

\textbf{Stage 2: Requirement-Level Semantic Evaluation.} Given the Game Specification, the corresponding Evaluation Requirements, and the final source code, the evaluator judges each requirement and provides supporting evidence. The evaluation focuses on whether the relevant code implements the behavior described by the requirement and whether the necessary inputs, state updates, system coordination, and outputs form a traceable execution path in the source code, rather than judging solely from function names, comments, or placeholder implementations.

Completeness Requirements and Richness Requirements are aggregated separately to measure the implementation of explicit functional requirements and hidden, deeper requirements. If an artifact fails the static gate or is missing, all its public and hidden requirement scores are set to zero. Because the GPT-6-astra combination uses the same model as the evaluator, its results constitute same-model self-evaluation, a setting in which LLM evaluators may favor their own generations~\citep{panickssery2024selfpref}. Evaluator configurations and evaluation templates are provided in Appendix E, and the final aggregation rules are given in Appendix E.4.

\subsection{Human Agreement}

We conduct a human cross-check on generated artifacts from a subset of player tasks, comparing human and automated judgments for the same Evaluation Requirements. Table~\ref{tab:main-4} presents the requirement-level confusion matrix.

\begin{table}[H]
\caption{Requirement-level human–machine confusion matrix. Only records that humans could judge and that were aligned requirement by requirement with the Astra evaluation are included. Judgment records for artifacts that failed the static gate are retained. Each artifact was assessed by a single human annotator.}
\label{tab:main-4}\centering\small
\begin{tabularx}{\linewidth}{*{3}{L}}\toprule
Human / machine & Fail (0) & Pass (1) \\
\midrule
Satisfied & 81 & 102 \\
Not satisfied & 47 & 11 \\
\bottomrule\end{tabularx}\end{table}

The human satisfaction rate is 75.9\%, higher than the judge's pass rate of 46.9\% on the same items. The primary direction of disagreement consists of requirements that humans judged satisfied but the automated evaluator failed, indicating a clear difference between static source-code judgments and human observation. The human assessment is therefore used to identify sources of disagreement in automated evaluation. Although the judge is considerably stricter than human play-testing, it preserves the ordering of systems: among the eight systems with at least 20 paired items, the per-system human and judge pass rates are almost perfectly rank-correlated (Spearman's $\rho = 0.95$, Table~\ref{tab:human-by-model}; \citealp{spearman1904}). The sample scope, label processing, and requirement-alignment method are described in Appendix F.

\subsection{Main Results}

We first compare project-level gate outcomes, Completeness, and Richness across combinations, and then examine differences by task source, the eight game-development capabilities, requirement difficulty, and platform. All requirement-level results are computed over all 1,980 model–task records: all public and hidden requirements receive a score of 0 if the artifact fails the static gate or is missing; otherwise, the recorded requirement judgments are used. The gate-failure-or-missing-artifact rate is reported separately. All comparisons below characterize the joint performance of each model--harness combination on the current fixed task set. The distinction between the fixed task universe and gate-passing common support is detailed in Appendix G.4.

\subsubsection{Overall Performance}

Table~\ref{tab:main-5} summarizes the results for each model–harness combination. Overall, 19.9\% of generation records are marked as static gate failures or missing artifacts. GLM-5.2 and DeepSeek-V4-Flash-0731 have the highest rate (26.7\%), and GPT-6-astra the lowest (13.3\%). These static gate outcomes indicate a substantial project-delivery gap beyond the implementation of specific functions: consistently delivering artifacts that satisfy the prerequisites for evaluation. The static review identifies delivery gaps in the entry artifacts and execution paths required for subsequent evaluation. Terminal generation statuses and their relationship to static gate outcomes are detailed in Appendix D.4.

Overall Completeness is 33.2\%, and Richness is 32.5\%. GPT-6-astra achieves the highest value on both metrics, but because this combination uses the same model as the evaluator, its results must be interpreted in light of the self-evaluation setting. After excluding this combination, the two pass rates are 30.8\% and 31.0\%, respectively. Among non-self-evaluated combinations, Fable 5.1 has the highest Completeness point estimate (44.7\%), while Opus 5 has the highest Richness point estimate (38.5\%), followed by Fable 5.1 (38.2\%). No non-self-evaluated combination achieves a pass rate above 50\% for either public or hidden requirements, leaving a substantial gap between complete game generation and the reliable implementation of a specification and its deeper system relationships.

Across the 11 model--harness combinations, the gate-failure-or-missing-artifact rate is strongly negatively associated with both Completeness and Richness (Spearman's $\rho \approx -0.70$ and $-0.72$, respectively). GPT-5.5 is an exception to this overall pattern: its gate-failure-or-missing-artifact rate is 15.6\%, the lowest among the non-self-evaluated combinations, but its Completeness and Richness are only 26.9\% and 27.9\%. This combination illustrates two distinct diagnostic dimensions: satisfying project-level evaluation prerequisites and implementing the specified requirements. The overall gate-failure-or-missing-artifact rate of 19.9\% also indicates that reliable delivery of complete game projects suitable for evaluation remains an unresolved challenge for the current model--harness combinations.

\begin{table}[H]
\caption{Main results for the model–harness combinations. CC denotes Claude Code and CX denotes Codex CLI; \textdagger{} marks same-model self-evaluation. Each combination is evaluated on 180 tasks, and requirement pass rates use all-record analysis.}
\label{tab:main-5}\centering\small
\begin{tabularx}{\linewidth}{@{}>{\hsize=1.65\hsize}L >{\hsize=.55\hsize}L >{\hsize=.8\hsize}L >{\hsize=1.2\hsize}L >{\hsize=.9\hsize}L >{\hsize=.9\hsize}L@{}}\toprule
Model & Harness & Generation records & Gate failure or missing artifact & Completeness & Richness \\
\midrule
DeepSeek-V4.1-Flash & CC & 180 & 20.6\% & 25.9\% & 27.6\% \\
DeepSeek-V4-Flash-0731 & CC & 180 & 26.7\% & 23.5\% & 26.9\% \\
Fable 5 & CC & 180 & 21.1\% & 39.0\% & 36.4\% \\
Fable 5.1 & CC & 180 & 17.8\% & 44.7\% & 38.2\% \\
GLM-5.2 & CC & 180 & 26.7\% & 24.3\% & 25.8\% \\
GLM-5.3 & CC & 180 & 21.7\% & 31.1\% & 31.8\% \\
Opus 5 & CC & 180 & 16.1\% & 37.2\% & 38.5\% \\
DeepSeek-V4-Pro & CX & 180 & 22.8\% & 22.2\% & 25.3\% \\
GPT-5.5 & CX & 180 & 15.6\% & 26.9\% & 27.9\% \\
GPT-5.6-sol & CX & 180 & 16.7\% & 33.4\% & 31.5\% \\
GPT-6-astra\textdagger{} & CX & 180 & 13.3\% & 57.1\% & 47.6\% \\
Overall & — & 1,980 & 19.9\% & 33.2\% & 32.5\% \\
\bottomrule\end{tabularx}\end{table}

\subsubsection{Completeness and Richness across Task Sources}

\begin{figure}[t]\centering\includegraphics[width=0.78\linewidth]{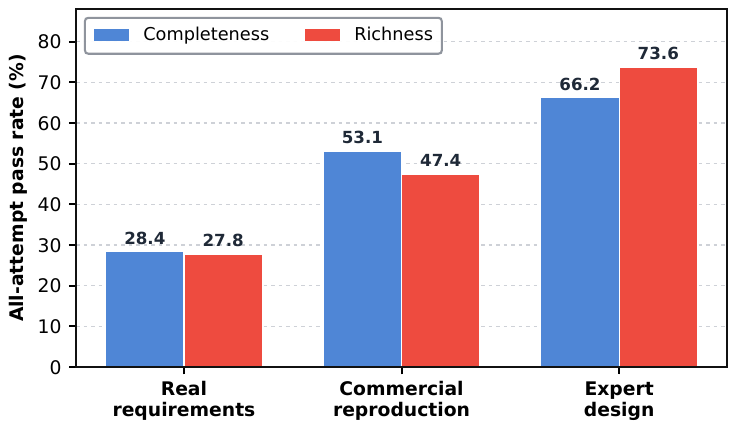}\caption{Completeness and Richness by task source}\label{fig:F06_source_heterogeneity}\end{figure}

Figure~\ref{fig:F06_source_heterogeneity} presents Completeness and Richness for real player requests, commercial-mechanics reconstruction, and expert-designed tasks.

For real player requests, Completeness is 28.4\% and Richness is 27.8\%; the corresponding values are 53.1\% and 47.4\% for commercial-mechanics reconstruction and 66.2\% and 73.6\% for expert-designed tasks. The current task set thus exhibits clear stratification by source: model–harness combinations have the lowest requirement pass rates on real player requests and higher rates on reconstruction and expert-designed tasks. This result shows that reporting only an overall mean across sources would conceal large performance differences across forms of requirements.

The relative ordering of public and hidden requirements also reverses across sources: Richness is higher for expert-designed tasks, whereas Completeness is higher for real-request and commercial-mechanics reconstruction tasks. The lower-performing requirement layer therefore varies by task source. Completeness and Richness capture different types of implementation gaps and should be reported separately rather than combined into a single aggregate score. The three sources consist of different task content and capabilities and serve as complementary diagnostic slices. Reporting both layers by source reveals whether publicly specified functionality or hidden system relationships remain less well implemented in each task category, a distinction that an overall mean would conceal.

\subsubsection{Capability-Level Performance}

\begin{figure}[t]\centering\includegraphics[width=0.88\linewidth]{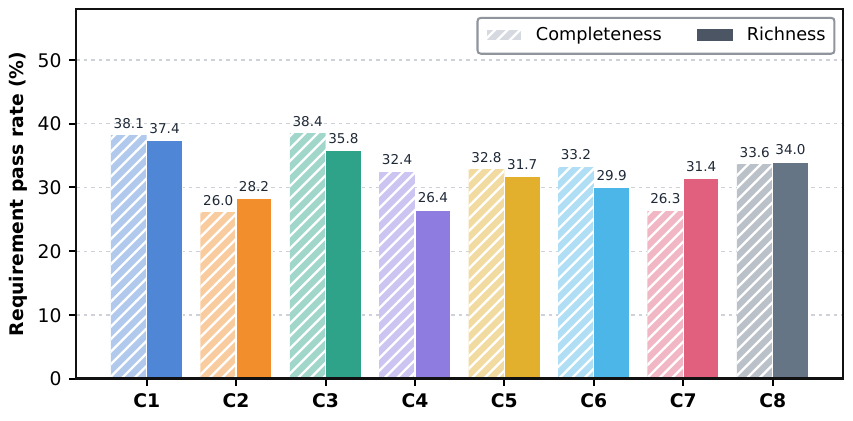}\caption{Requirement pass-rate point estimates by capability. Coverage and aggregation sensitivity are reported in Appendix G.2.}\label{fig:F03_capability}\end{figure}

Figure~\ref{fig:F03_capability} aggregates Completeness and Richness according to each requirement's primary capability assignment, showing implementation performance across the eight game-development capabilities.

C1 Core Rules and C3 Numerical Design have the highest combined pass-rate point estimates, at 37.8\% and 37.4\%, respectively, while C2 AI/NPC and C7 Feedback and Game Feel have the lowest, at 27.0\% and 28.1\%. By layer, Completeness has its highest point estimate for C3 (38.4\%) and lowest for C2 (26.0\%), close to C7 (26.3\%). Richness has its highest point estimate for C1 (37.4\%) and its lowest for C4 Interaction and Input (26.4\%), based on 64 hidden requirements. Taken together, this pattern indicates that current systems more readily implement inspectable rules and numerical relationships, while autonomous NPC behavior and feedback and game-feel mechanisms remain more prominent implementation gaps. Improving complete game generation therefore requires strengthening the behavior and feedback loops that connect game states to the player experience, alongside implementing core gameplay.

The capability profiles for Completeness and Richness differ, showing why capability diagnosis should retain the distinction between public and hidden requirements. The overall pattern of higher combined pass rates for C1/C3 than C2/C7 persists under multilabel aggregation and after excluding same-model self-evaluation. All eight capabilities have higher point estimates for Mobile than for PC, but cross-track comparisons also reflect differences in task content and protocols (Appendices A.5 and E.1). The persistence of the C1/C3 versus C2/C7 pattern across aggregation protocols supports a consistent improvement priority: strengthening autonomous NPC behavior and feedback and game-feel mechanisms. Capability coverage and multilabel/no-self-evaluation sensitivity analyses are provided in Appendix G.2.

\subsubsection{Performance across Difficulty}

\begin{figure}[t]\centering\includegraphics[width=0.62\linewidth]{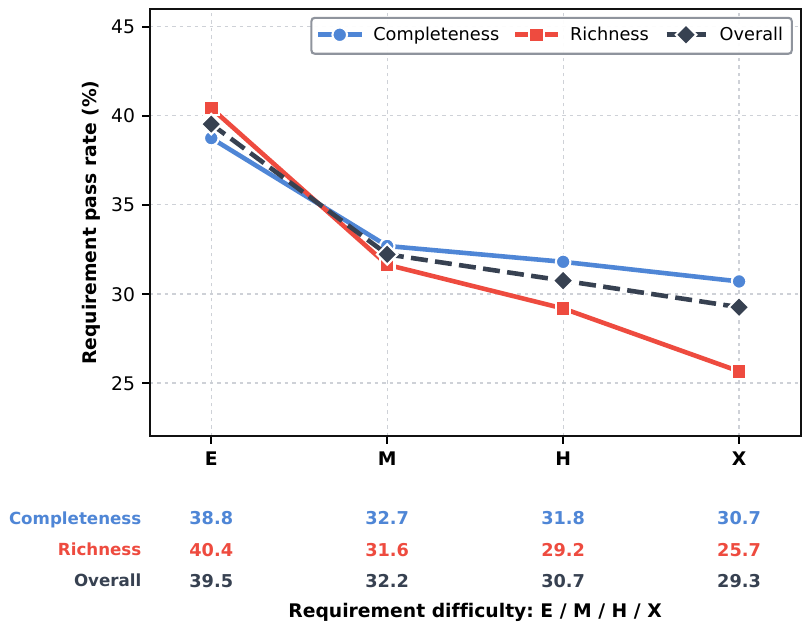}\caption{Pass rates across requirement-difficulty strata. E, M, H, and X denote Easy, Medium, Hard, and Extreme, respectively. The three curves show Completeness, Richness, and the combined pass rate for both requirement types. Difficulty strata follow the equal-count partition defined in Appendix C. These are pooled descriptive curves; subset-specific rates and composition sensitivity are reported in Appendix G.1.}\label{fig:F04_requirement_difficulty}\end{figure}

Figure~\ref{fig:F04_requirement_difficulty} presents the four requirement-difficulty strata and their overall pass rates. Each stratum contains 450 requirements and 4,950 model–requirement records.

The overall requirement pass rate decreases from 39.5\% for Easy to 32.2\% for Medium, 30.7\% for Hard, and 29.3\% for Extreme, a difference of 10.3 percentage points between Easy and Extreme. Completeness decreases monotonically from 38.8\% to 30.7\%, and Richness from 40.4\% to 25.7\%. Overall, difficulty is weakly negatively correlated with the 11-model average pass rate for each requirement (Spearman's $\rho = -0.1201$). The pooled decline is sensitive to subset composition, and within-subset associations are weak. Requirement-difficulty strata therefore describe an aggregate performance gradient in this benchmark, while interpretation within a particular source or track requires its own requirement composition. Full results and composition sensitivity are provided in Appendix G.1.

\subsubsection{Performance across Platforms}

\begin{figure}[t]\centering\includegraphics[width=0.88\linewidth]{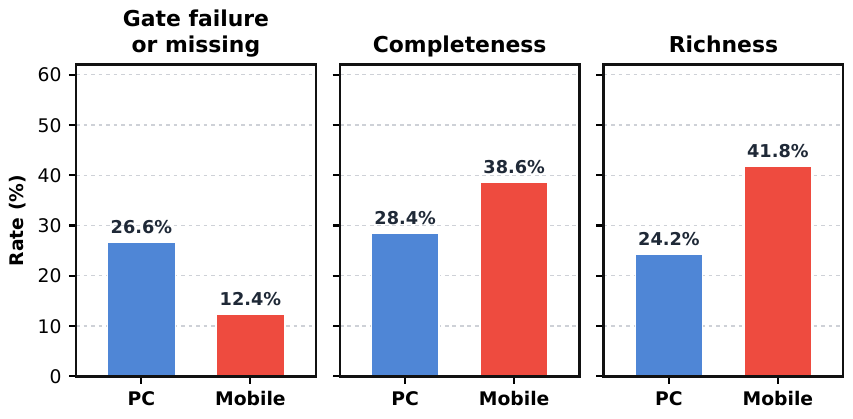}\caption{Results by platform}\label{fig:F05_platform}\end{figure}

Figure~\ref{fig:F05_platform} compares the gate-failure-or-missing-artifact rate, Completeness, and Richness between the PC Web and Mobile Web task subsets.

The gate-failure-or-missing-artifact rate is lower for Mobile than for PC (12.4\% vs. 26.6\%). Completeness/Richness is 28.4\%/24.2\% for PC and 38.6\%/41.8\% for Mobile. The direction is consistent across all 11 model--harness combinations and all eight capabilities: both requirement pass rates are higher on Mobile for every combination, and the combined capability pass rates follow the same direction.

After excluding the 10 expert-designed tasks unique to PC, both platforms have the same source composition: 75 player tasks and 10 commercial-mechanics reconstruction tasks. On this shared-source support set, the gate-failure-or-missing-artifact rates for PC and Mobile are 29.4\% and 12.4\%, Completeness is 23.9\% and 38.6\%, and Richness is 18.4\% and 41.8\%, respectively; the direction of the difference remains unchanged. With matched source proportions, Mobile retains higher Completeness and Richness by 14.7 and 23.3 percentage points, respectively. The platform-specific task tracks therefore show a consistent implementation gap under both the full-task and shared-source analyses. Task content and evaluation protocols for these tracks are described in Appendices A.5 and E.1.

The consistency of this direction across all 11 combinations and eight capabilities shows that the platform-specific task tracks reveal systematic differences that an overall leaderboard cannot show. In the current experiment, Mobile combines a lower static gate-failure-or-missing-artifact rate with higher requirement pass rates, indicating stronger project delivery as measured by the static gate and more complete requirement implementation on this task track. Reporting platform-specific results therefore helps distinguish gaps in project delivery from gaps in requirement implementation. The two tracks differ in task content, requirement adaptation, and judge-call structure (Appendices A.5, B.5, and E.1); the shared-source sensitivity analysis is reported in Appendix G.3.

\subsection{Case Analysis}

We examine two artifacts generated by Fable 5.1, the 2D industrial simulation \texttt{hard\_city\_2d\_009} and the 3D puzzle game \texttt{online\_l5\_puzzle\_3d\_0223}, to connect requirement-level judgments to specific implementation issues. The cases focus on gaps between functional coverage and requirement satisfaction, particularly in cross-system coordination and state consistency.

\textbf{2D industrial simulation.} Static review passes three of six Completeness Requirements and one of four Richness Requirements. The game implements interconnected conveyor, fluid, and power networks, but the evaluation identifies failures in shared resource constraints, production-state handling, and simulation timing.

One failure concerns fluid-network capacity. The implementation assigns capacity limits independently to adjacent edges without enforcing a shared throughput limit at pipe nodes. According to the evaluator's rationale, a pipe with a nominal capacity of 3 L/s may simultaneously carry two flows of 3 L/s, exceeding its shared capacity. This issue affects both a Completeness Requirement concerning capacity-constrained fluid delivery and a Richness Requirement concerning network loops and branch contention. Although local capacity constraints are implemented, they do not guarantee the required system-level behavior.

Another failure concerns resource accounting during structural changes. The implementation supports material transfers, recipe-based consumption, and network reconstruction after entity removal. However, the static review identifies an accounting gap when a machine is removed during an active production cycle after its input materials have been consumed. Updating network connectivity does not ensure that resources committed to unfinished production are consistently accounted for.

The static review also identifies a timing issue. Although the simulation uses fixed time steps and seeded randomness, the rendering loop caps elapsed frame time before accumulating simulation time. As a result, long frame intervals may discard elapsed time, causing simulation progression to depend on rendering conditions rather than the intended elapsed duration.

\textbf{3D puzzle game.} Static review passes five of six Completeness Requirements but only one of four Richness Requirements. One failed hidden requirement concerns the interaction between escape and enemy capture. According to the evaluator's rationale, the exit-input callback immediately commits the escape state, after which the game update skips enemy logic because a terminal state has already been set. If exit interaction and capture conditions coincide within an update frame, this path may resolve escape before capture, violating the required capture-failure priority. Both mechanisms are implemented, but their handling of the shared terminal state does not enforce the required resolution order.

These cases illustrate a gap between functional coverage and complete requirement satisfaction. Even when relevant mechanisms are implemented, shared resource constraints, production-state transitions, and event ordering may remain incorrect. Requirement-level static evaluation helps localize these implementation gaps by tracing specific requirements to source-code evidence, revealing issues that aggregate scores alone cannot distinguish.

\section{Conclusion}

This paper presents OCTAVE, which decomposes complete game generation into capability coverage and requirement-level implementation. The benchmark draws tasks from real player requests, commercial-mechanics reconstruction, and human expert design; distinguishes PC Web from Mobile Web during construction; and uses eight Capability Dimensions to organize task labels and requirement assignments. Completeness and Richness Requirements further translate explicit functional implementation and hidden system relationships into inspectable source-code requirements.

Experiments with 11 model–harness combinations show clear gaps in both project delivery and requirement implementation in current systems. Completeness is 33.2\%, Richness is 32.5\%, and the gate-failure-or-missing-artifact rate is 19.9\%. After excluding same-model self-evaluation, even the best-performing combination has pass rates below 50\% for both public and hidden requirements. Across combinations, the gate-failure-or-missing-artifact rate is strongly negatively associated with both Completeness and Richness (Spearman's $\rho \approx -0.70$ and $-0.72$, respectively). GPT-5.5 is an exception, with a low gate-failure-or-missing-artifact rate but also low requirement pass rates, highlighting the need to diagnose project-delivery prerequisites and requirement implementation as separate dimensions.

Requirement-level decomposition further reveals structure that aggregate scores cannot capture. The lower-performing requirement layer varies by task source, making separate reporting of Completeness and Richness necessary for locating implementation gaps. Combined pass rates are relatively higher for C1 Core Rules and C3 Numerical Design and lower for C2 AI/NPC and C7 Feedback and Game Feel; this recurring pattern identifies autonomous behavior and feedback mechanisms as priorities for improvement. Requirement-difficulty strata reveal an aggregate performance gradient, while weak within-subset associations and composition sensitivity show the need to interpret difficulty jointly with task source and track. Mobile has higher requirement pass rates across all combinations and eight capabilities in the current task set, demonstrating the value of platform-specific diagnosis beyond an overall leaderboard; differences in task content and protocols remain. In the human assessment of a limited sample, the principal disagreement consists of requirements judged satisfied by humans but failed by the automated evaluator, further showing the value of requirement-level evidence for identifying differences between static judgments and actual experience. On paired items the judge is about 29 percentage points stricter than human play-testing while preserving system order (Spearman's $\rho = 0.95$), so absolute pass rates are best read as a lower bound rather than as the share of features that work for a player.

The core value of OCTAVE lies in providing a diagnostic chain from task capabilities through requirement implementation to aggregate results. It can show which requirements receive inspectable implementations from an agent, which capability dimensions still exhibit gaps, and how those results are affected by task source and the evaluation support set. Future work should combine execution traces, evidence of player experience, and representative artifact cases to further validate these requirement-level diagnoses and analyze reproducible, systematic failure patterns.

\FloatBarrier
\appendix
\counterwithin{figure}{section}
\counterwithin{table}{section}

The appendices provide supplementary details on task construction, requirement development, complexity scoring, generation configurations, automated evaluation, the human cross-check, and statistical analyses. Released materials and dataset documentation are listed in Appendix H.

\section{Task Sources, Capability-Driven Selection, and Data Distribution}

\subsection{Extraction, Cleaning, and Provenance of Real Player Requests}

This section describes how raw player requests were transformed into traceable candidate tasks suitable for selection.

The source material consists of the first prompts in game-generation requests. In the first extraction, 1,000 entries were collected under quotas of 400 ceiling, 500 mid, and 100 floor, comprising 617 2D entries and 383 3D entries. Another batch divided the source material into five slices for preliminary screening and produced 151 candidates. Rewriting, merging, and deduplication across batches produced the player candidate pool, from which entries were included after review.

The construction rules filter out requests that are not games, lack sufficient information, duplicate other content, or cannot be implemented in the target setting. They rewrite themes, core loops, and system relationships into judgeable specifications. The rewriting preserves thematic characteristics while adding system states, causal relationships, resource constraints, win/loss resolution, and behavioral boundaries. The resulting descriptions are then normalized according to the construction rules.

The main player candidate pool contains 250 tasks. The 150 selected tasks comprise 11 from the \texttt{hard} source family, 48 from \texttt{hard\_rewrite}, 19 from \texttt{online\_l}, and 72 from \texttt{v3}. Family identifiers support source tracing. Provenance is linked through source tables, initial candidates, and production-version identifiers; some early entries retain only within-batch identifiers. Statistics for the final list are provided in this section and in Appendix A.4.

\subsection{Commercial-Mechanics Reconstruction}

This section explains how public game mechanics were transformed into independent benchmark tasks and distinguishes facts about the original games from supplementary benchmark rules.

Reconstruction is based on public store pages, developer descriptions, guides, Wikis, and reviews. Researchers proposed candidates, and LLMs assisted with retrieval and construction. This process yielded 31 drafts (16 Steam and 15 iOS), of which 20 were retained after structured review (10 each for PC and Mobile).

The review examined system coupling, accessory systems, coverage and overreach of public requirements, the information content of requirement slots, judgeability, thresholds and verification conventions, ambiguous or universal statements, inferability of hidden requirements, data contamination, and evidence sources. Seven PC tasks and eight Mobile tasks with a score of 4 were retained first; the remaining tasks filled the platform quotas based on defect reparability, structure, and evidence. Evidence labels are STORE, DEV, WIKI, GUIDE, REVIEW, and INFER.

\subsection{Expert-Designed Tasks}

This section describes the scope of expert-designed tasks and their relationship to the player-task selection process.

Experts designed 10 final tasks, all for PC Web. Each task contained six public requirements and four hidden requirements beginning at the design stage. These tasks supplement multisystem relationships and computational and engineering constraints. They were added directly to the final set and did not participate in the 250-to-150 player-task selection.

\subsection{Capability Coverage Before and After Selection and in the Final Set}

We first annotated the 250 player candidate tasks using the C1–C8 capability framework and then selected 150 tasks primarily according to the balance of their capability coverage. The following analysis reports multilabel capability coverage before and after selection. Commercial-mechanics reconstruction and expert-designed tasks were incorporated directly into the final set according to their respective construction protocols.

Table~\ref{tab:coverage} lists the number and proportion of tasks covering each capability in the player candidate set and the selected set. Because tasks have multilabel annotations, a single task may be counted under multiple Capability Dimensions.

\begin{table}[H]
\caption{Player-task capability coverage before and after selection.}
\label{tab:coverage}\centering\small
\begin{tabularx}{\linewidth}{*{3}{L}}\toprule
Paper capability & Coverage in 250 player candidates & Coverage in 150 selected player tasks \\
\midrule
C1 & 250 (100.0\%) & 150 (100.0\%) \\
C2 & 215 (86.0\%) & 133 (88.7\%) \\
C3 & 224 (89.6\%) & 130 (86.7\%) \\
C4 & 115 (46.0\%) & 74 (49.3\%) \\
C5 & 154 (61.6\%) & 91 (60.7\%) \\
C6 & 111 (44.4\%) & 74 (49.3\%) \\
C7 & 56 (22.4\%) & 31 (20.7\%) \\
C8 & 101 (40.4\%) & 80 (53.3\%) \\
\bottomrule\end{tabularx}\end{table}

We use Jain's index \cite{jain1984fairness} to measure the evenness of coverage across the eight Capability Dimensions. Let $c_i$ denote the number of tasks in a set that cover capability $C_i$. Then

\[
J(c)=\frac{\left(\sum_{i=1}^{8}c_i\right)^2}{8\sum_{i=1}^{8}c_i^2}.
\]

\begin{figure}[H]\centering\includegraphics[width=0.95\linewidth]{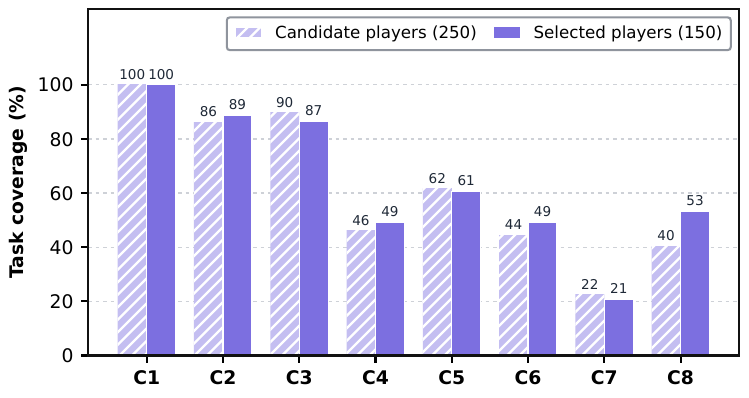}\caption{Capability coverage before and after player-task selection}\label{fig:F02_candidate_coverage_retrospective}\end{figure}

Jain's index is 0.8482 for the player candidate set and 0.8696 for the selected set; the average number of labels per task is 4.904 and 5.087, respectively. The selection therefore increases the measured evenness of capability coverage. Figure~\ref{fig:F02_candidate_coverage_retrospective} compares capability coverage before and after selection.

The 100 excluded tasks contain 463 labels in total, averaging 4.630 labels per task, with a Jain's index of 0.7964.

The player-task selection retained 150 of 250 candidates using capability coverage as the primary selection criterion. The remaining 30 reconstruction and expert-designed tasks were incorporated directly into the final set. The selection table does not preserve task-specific exclusion rationales for the 100 excluded tasks; it reports capability coverage at the set level before and after selection.

For the full final set of 180 tasks, including player, reconstruction, and expert-designed tasks, the capability labels total 878, with a Jain's evenness index of 0.8573. Each mutually exclusive task grouping sums to 180. The primary capabilities of public and hidden requirements are counted according to the task and requirement definitions used in this paper.

\subsection{Platform Composition and Task Track Separation}

This section disaggregates final tasks by platform and task source. Because all expert-designed tasks are on PC Web, the source compositions of PC and Mobile differ.

\begin{table}[H]
\caption{Task-source composition by platform.}
\label{tab:platform}\centering\small
\begin{tabularx}{\linewidth}{*{5}{L}}\toprule
Platform & Player tasks & Reconstruction tasks & Expert-designed tasks & Total \\
\midrule
PC Web & 75 & 10 & 10 & 95 \\
Mobile Web & 75 & 10 & 0 & 85 \\
Total & 150 & 20 & 10 & 180 \\
\bottomrule\end{tabularx}\end{table}

Both platform tracks follow the same sequence of task and requirement construction, review, generation, and static scoring; requirements are determined before generation. Mobile Web tasks target mobile browsers and have separately constructed touch-input and platform constraints. The shared-source sensitivity analysis is provided in Appendix G.3.

\section{Specification and Requirement Construction and Expert Review}

\subsection{Specification Structure and Implementation Freedom}

A specification describes a task's objectives, mechanics, and required behaviors from the perspectives of the player and game design. Player tasks use the following four-part template.

\begin{table}[H]
\caption{Four-part template for player-task specifications.}
\label{tab:specification}\centering\small
\begin{tabularx}{\linewidth}{@{}>{\hsize=.75\hsize}L >{\hsize=1.25\hsize}L@{}}\toprule
Section & Content \\
\midrule
\texttt{[Game]} & Game genre and core concept. \\
\texttt{[Core Mechanics]} & Principal systems, states, causal relationships, and gameplay loop. \\
\texttt{[Must Implement]} & Required behaviors that support the core design. \\
\texttt{[Creative Freedom]} & Presentation, content, and extensions left to the implementer. \\
\bottomrule\end{tabularx}\end{table}

Mobile tasks may add platform-specific interaction requirements to this structure. Content under [Creative Freedom] remains an implementation choice and does not introduce additional mandatory conditions. The final task descriptions from different sources are governed by their respective frozen texts and may vary in length according to task content.

Specifications do not prescribe fixed APIs, function names, code structures, or test interfaces. Determinism, conservation, and scale constraints are expressed as behavioral requirements, allowing behavior-level judgments while preserving freedom of engineering implementation.

\subsection{Construction of Public Requirements}

Each task contains six public Completeness Requirements that define the conditions for completing its core design. Requirements should describe identifiable preconditions, actions, state changes, and expected outcomes. The construction framework for player tasks covers single-system correctness, cross-system coupling, win/loss or loop resolution, and determinism, scale, or boundary constraints; the specific order varies by task. The object of judgment is actual behavior, not the presence of particular code identifiers.

Requirements may use formulations based on state consistency, comparison, temporal ordering, statistics, conservation, convergence, or hashing, while avoiding inexhaustible universal claims, unsupported thresholds, and subjective adjectives. Automated judgments follow the static source-code review process in Appendix E.1.

\subsection{Construction of Hidden Requirements}

Each task additionally contains four hidden Richness Requirements that are not provided to the generation agent. These requirements examine edge cases, state consistency, second-order relationships, or cross-system coupling naturally inferred from the task description. They should target behavior that goes beyond direct implementation of public requirements but is still implied by the existing design, without introducing unrelated new systems.

Examples mapping public to hidden requirements are provided in Appendix B.4. Released task files and dataset documentation are listed in Appendix H.

\subsection{Examples of Public and Hidden Requirements}

The following example is drawn from the frozen PC task \texttt{hard\_action\_rpg\_3d\_017} and illustrates how the two requirement layers examine the same mechanic. The task description states that "after a wall is destroyed, both the player and enemies update their routes." The corresponding requirements are:

\par\medskip
\noindent\textbf{Completeness}\par
After a wall connecting rooms is destroyed, blockage, collision, and navigation update synchronously. For the same start and destination, the player changes from being unable to pass through the wall to being able to traverse the opening, while enemies that previously detoured or could not reach the player recalculate their paths and use the opening when the new route makes the player reachable.\par\medskip
\noindent\textbf{Richness}\par
When walls are destroyed in succession or a wall is destroyed while an enemy is moving along an old route, affected paths must update. Intact walls must continue to block movement, openings must retain no residual collision, temporarily unreachable enemies must reroute or wait, and pathfinding must resume when a route becomes available again.\par\medskip

The public requirement explicitly specifies the before-and-after comparison, state updates, and observable outcomes; the hidden requirement changes event timing and reachability conditions within the same system. Both permit different navigation implementations. The task description can be located by \texttt{task\_id} in the \href{https://github.com/rezona-ai/octave-bench/blob/main/data/octave\_180\_en.jsonl}{released English task file}.

\subsection{Revisions to Mobile Web Requirements}

For the 75 player-request Mobile tasks (Mobile-75), the requirement-construction records contain 450 public requirements and 300 hidden requirements. The following table distinguishes full replacement, same-source revision, and retention without change.

\begin{table}[H]
\caption{Revisions to Mobile Web requirements.}
\label{tab:revisions}\centering\small
\begin{tabularx}{\linewidth}{*{5}{L}}\toprule
Layer & Full replacement & Same-source revision & Retained unchanged & Total \\
\midrule
Public & 196 (43.6\%) & 73 (16.2\%) & 181 (40.2\%) & 450 \\
Hidden & 99 (33.0\%) & 29 (9.7\%) & 172 (57.3\%) & 300 \\
\bottomrule\end{tabularx}\end{table}

Changes to public requirements affect all 75 tasks, and changes to hidden requirements affect 62 tasks. Revisions focus on touch-input paths, tolerance at boundaries, and requirement judgeability.

\subsection{Expert Composition and Review Scope}

The construction-review group comprised 12 experts: six developers, three designers, and three testers. Their mean experience was 8.7 years (range: 5–15 years).

Beginning at the design stage for the final tasks, experts participated in construction and review using six public and four hidden requirements per task. The review covered clarity of task descriptions, feasibility, system relationships, capability mappings, interaction paths, and complexity. Recorded batches include requirement review for 75 Mobile tasks, capability-label review for 250 tasks, and construction review for the 10 expert-designed tasks.

\subsection{Review Records and Revision Feedback}

The retained materials include methodological descriptions, final tasks, a classification summary of Mobile revisions, and capability-label verification results. Mobile revision counts are provided in Appendix B.5.

Supplementary feedback from external evaluators constitutes a separate form of evidence. It covers 23 tasks, 312 human scores, and 288 quality-control notes and identified 32 rubric issues in 13 tasks, leading to 26 rewrites and six splits. This feedback was used to revise observability, scoring scope, and compound requirements.

\section{Design-Complexity Scoring and Stratification}

\subsection{Complexity Dimensions and Scoring Anchors}

Design complexity describes the complexity of system design, algorithmic implementation, and coupling semantics required to satisfy the task description and requirements. It is assessed during construction independently of model pass rates. We use six dimensions, each assigned an ordinal score from 0 to 3 according to the following anchors.

\begin{table}[H]
\caption{Design-complexity dimensions and scoring anchors.}
\label{tab:complexity}\centering\small
\begin{tabularx}{\linewidth}{*{5}{L}}\toprule
Dimension & 0 & 1 & 2 & 3 \\
\midrule
D1 Breadth of system coupling & A single loop and accessory elements & Two hard systems, primarily with one-way relationships & Three hard systems, including bidirectional relationships & At least four densely coupled hard systems or a causal chain spanning three systems \\
D2 Algorithmic and simulation core & State machines and numerical tables & One standard algorithm & A nontrivial core & Multiple jointly operating cores or a simulation-level core \\
D3 Verification rigor & Functional presence & Seed consistency or single-point reproducibility & Comparison or conservation & Replay/hashing combined with multiple types of constraints \\
D4 Dynamic governance and adaptation & No dynamic balancing & A single feedback loop & Multiple feedback loops or persistent stability constraints & Closed-loop adaptation and interlocking multiple loops \\
D5 Multi-agent and social cognition & None, or scripted NPCs & State machines and simple perception & Local information or opponent modeling & Multi-party games, reputation, or shared beliefs \\
D6 Scale and performance & No scale requirement & A soft stability constraint & Explicitly tens to hundreds of entities & Thousands of entities or hundreds of nodes, with hard smoothness requirements \\
\bottomrule\end{tabularx}\end{table}

Here, a hard system is a system with persistent state, update rules, and independently describable behavior. A score display or purely artistic accessory does not count toward the breadth of system coupling. The six dimensions characterize the constraints imposed by each requirement.

\subsection{Requirement Scoring and Level Assignment}

The difficulty score for each requirement is defined as $a_{DS} = a_{D1} + a_{D2} + 0.8a_{D3} + 0.8a_{D4} + 0.7a_{D5} + 0.5a_{D6}$. Requirements are then sorted by $a_{DS}$, \texttt{task\_id}, requirement layer, and within-layer index and divided by position into four equal-count strata: Easy, Medium, Hard, and Extreme. The stratum boundaries are 1.7, 2.6, and 3.6, with 450 requirements per stratum. When identical scores cross a boundary, assignment is determined by the stable ordering above.

\section{Generation Configurations and Process Records}

\subsection{Model and Agent-Harness Configurations}

The experimental subjects comprise 11 combinations of base models and agent harnesses. The table below lists the model name, agent harness and version, and reasoning-effort setting. Claude Code~\citep{claudecode2025} is version 2.1.263, and Codex CLI~\citep{codexcli2025,codexharness2026} is version 0.147.0. Reasoning-effort settings are not equivalent across models. "None" in the table indicates that this setting was not specified; it does not imply zero reasoning expenditure.

\begin{table}[H]
\caption{Model and agent-harness configurations.}
\label{tab:models}\centering\small
\begin{tabularx}{\linewidth}{@{}>{\hsize=1.65\hsize}L >{\hsize=1.5\hsize}L >{\hsize=.85\hsize}L >{\hsize=.55\hsize}L >{\hsize=.45\hsize}L@{}}\toprule
model\_key & Display name & Agent harness & Version & effort \\
\midrule
\texttt{cc\_\allowbreak deepseek\_\allowbreak flash\_\allowbreak 41} & DeepSeek-V4.1-Flash (API id deepseek-flash) & Claude Code & 2.1.263 & None \\
\texttt{cc\_\allowbreak deepseek\_\allowbreak v4\_\allowbreak flash} & DeepSeek-V4-Flash-0731 & Claude Code & 2.1.263 & None \\
\texttt{cc\_\allowbreak fable\_\allowbreak 5} & Fable 5 xhigh & Claude Code & 2.1.263 & xhigh \\
\texttt{cc\_\allowbreak fable\_\allowbreak 5\_\allowbreak 1} & Fable 5.1 xhigh & Claude Code & 2.1.263 & xhigh \\
\texttt{cc\_\allowbreak glm\_\allowbreak 5\_\allowbreak 2} & GLM-5.2 & Claude Code & 2.1.263 & None \\
\texttt{cc\_\allowbreak glm\_\allowbreak 5\_\allowbreak 3} & GLM-5.3 & Claude Code & 2.1.263 & None \\
\texttt{cc\_\allowbreak opus\_\allowbreak 5} & Opus 5 xhigh & Claude Code & 2.1.263 & xhigh \\
\texttt{codex\_\allowbreak deepseek\_\allowbreak v4\_\allowbreak pro} & DeepSeek-V4-Pro & Codex CLI & 0.147.0 & high \\
\texttt{codex\_\allowbreak gpt\_\allowbreak 5\_\allowbreak 5} & GPT-5.5 high & Codex CLI & 0.147.0 & high \\
\texttt{codex\_\allowbreak gpt\_\allowbreak 5\_\allowbreak 6\_\allowbreak sol} & GPT-5.6-sol high & Codex CLI & 0.147.0 & high \\
\texttt{codex\_\allowbreak gpt\_\allowbreak 6\_\allowbreak astra} & GPT-6 astra high & Codex CLI & 0.147.0 & high \\
\bottomrule\end{tabularx}\end{table}

\subsection{Runtime Environment and Artifact Retention}

Each task begins in an independent workspace containing no preexisting game logic or task-implementation code. The agent autonomously installs dependencies, constructs the project, implements it, and debugs it. The final workspace, source code, and process records are retained. Evaluation operates on the final artifact and the final requirements. Harness \texttt{success} is determined according to build-delivery conditions, including conditions on the distribution entry point, and is not equivalent to successful execution observed in a standardized browser environment.

\subsection{Budget, Permission, and Retry Rules}

The generation budget for each task is 120 minutes. The agent may finish early; upon reaching the limit, execution stops and the final workspace is retained. Stopping and cleanup may cause observed elapsed time to slightly exceed 7,200 seconds, but this does not change the budget definition. Each task receives only one initialization instruction, with no additional human prompts during development. An empty workspace does not restrict the use of libraries or frameworks.

The harness permits shell-command execution and dependency installation and is configured to bypass interactive approval and sandbox restrictions. The research protocol does not uniformly disable network access; actual access depends on the runtime environment.

Only one final record is included for each model–task combination, yielding 1,980 records.

\subsection{Terminal Generation Statuses and Runtime Statistics}

Among the 1,980 generation records, the terminal statuses are 1,945 \texttt{success}, 26 \texttt{timeout}, four \texttt{agent\_\allowbreak error}, and five \texttt{refusal}. Among records marked as gate failures or missing artifacts, 375 have the \texttt{success} status, 10 have the \texttt{timeout} status, four have the \texttt{agent\_\allowbreak error} status, and five have the \texttt{refusal} status. Records without such a mark include 1,570 \texttt{success} and 16 \texttt{timeout}. These counts sum to the 1,980 records. \texttt{success}, \texttt{timeout}, \texttt{agent\_\allowbreak error}, \texttt{refusal}, and \texttt{no\_dist} are terminal statuses of the generation process; gate failure is a marker obtained by statically inspecting an artifact after generation. They belong to different stages and cannot substitute for each other.

Per-model runtime medians and quartiles are summarized in the released runtime records; cross-tabulation of terminal statuses and gate outcomes follows the gate-counting protocol below. The output-token field is zero for 28 records (26 \texttt{timeout} and two \texttt{agent\_\allowbreak error}); actual token consumption cannot be reliably determined from this field for these records. Terminal status and artifact completeness are assessed separately using status records and static evaluation, respectively. Runtime includes stopping and cleanup.

\section{Automated Evaluation and Score Aggregation}

\subsection{Static-Review Inputs and Evidence Requirements}

Inputs to automated evaluation include the specification, public and hidden requirements, source code, file list, and build information. Four templates cover PC Web and Mobile Web in both 2D and 3D, and all explicitly use static source-code review only. The evaluation does not receive screenshots, videos, browser logs, or interaction traces. A judgment must refer to files, functions, states, and data flows on actual execution paths; names, comments, the README, and dead code do not constitute evidence.

Evaluation proceeds in two stages. The first checks the entry point, dependencies, initialization, state containers, and loops to identify project-level blockers. The second gives a 0/1 judgment and an evidence-based rationale for each requirement. Judgment rationales for public and hidden requirements are stored separately. PC-75 uses a separate call for hidden requirements, whereas Mobile-75 and Final-30 evaluate both layers in the same call. For aggregation, all requirement scores are set to zero when the static gate fails or the artifact is missing (Appendix E.4). Auxiliary dimensions use a score from 0 to 5 and serve only to describe the feedback and scene layers; they do not contribute to Completeness, Richness, or scores for the eight capabilities.

\subsection{Evaluator Model and Invocation Interface}

The evaluator for this batch is \textbf{gpt-6-astra}~\citep{gpt6astra2026}, invoked primarily through the Responses API. Within the 75 player PC tasks, 15 judgment rationales for \texttt{cc\_deepseek\_flash\_41} were generated through Codex CLI (four public and 11 hidden, with \texttt{effort} set to \texttt{low}). In the main table, \texttt{codex\_gpt\_6\_astra} uses the same model as the evaluator and is marked as self-evaluation. A sensitivity analysis excluding this combination is provided in Appendix G.2.

\subsection{Evaluation Templates and Requirement-Judgment Rules}

The complete templates appear as string constants in four entry points: \href{https://github.com/rezona-ai/octave-bench/blob/main/llm-eval/2D\_eval.py}{PC 2D}, \href{https://github.com/rezona-ai/octave-bench/blob/main/llm-eval/3D\_eval.py}{PC 3D}, \href{https://github.com/rezona-ai/octave-bench/blob/main/llm-eval/2D\_mobile\_eval.py}{Mobile 2D}, and \href{https://github.com/rezona-ai/octave-bench/blob/main/llm-eval/3D\_mobile\_eval.py}{Mobile 3D}. The actual evaluator configuration is described in Appendix E.2.

The tool truncates inputs according to the source-code budget and marks omitted content. The retained records do not completely preserve the extent or frequency of truncation for each invocation.

\textbf{Requirement-judgment rules.} The evaluator distinguishes affirmative functionality, technical or asset constraints, and prohibitive requirements. A compound requirement receives a score of 1 only if all its subconditions are satisfied; no partial credit is awarded. Evidence must jointly include the logic definition, execution entry point, state consumption, and player-facing output. A dead branch unrelated to a requirement's causal path is noted only as a risk and does not independently cause the requirement to fail. The template uses the original requirement text as the result key, with the value storing a 0/1 score and a Chinese evidence-based rationale; hidden requirements are stored in \texttt{hidden\_probes}. The original structures of these output fields are preserved in the evaluation entry points.

\subsection{Gate Coding and All-Record Aggregation Formulas}

Let $\Omega$ denote all model–task records, $A_{mt}$ indicate that an artifact exists, and $G_{mt}$ indicate static gate failure. For PC player tasks, the public gate marker in the result package is used. For Mobile and expert/reconstruction tasks, $G_{mt}=1$ when the number of passed public requirements is zero and at least one public-requirement judgment rationale contains "unable to run." Missing artifacts are marked separately.

Let $\widetilde y^C_{mtj},\widetilde y^R_{mtj}\in\{0,1\}$ denote the recorded judgments for the six public and four hidden requirements of each task. The effective scores $y^C_{mtj},y^R_{mtj}$ used in aggregation are defined by the gate and artifact indicators introduced above:
\[
y^L_{mtj}=\mathbf{1}[A_{mt}=1\ \land\ G_{mt}=0]\,\widetilde y^L_{mtj},\qquad L\in\{C,R\}.
\]
Thus, every public and hidden requirement receives zero if the artifact fails the gate or is missing. The gate-failure-or-missing-artifact rate $B$, Completeness $S_C$, and Richness $S_R$ are defined as

\[
B=\frac{\sum_{(m,t)\in\Omega}\mathbf{1}[G_{mt}=1\ \lor\ A_{mt}=0]}{|\Omega|},\qquad
S_C=\frac{1}{|\Omega|}\sum_{(m,t)\in\Omega}\frac{\sum_{j=1}^{6}y^C_{mtj}}{6},\qquad
S_R=\frac{1}{|\Omega|}\sum_{(m,t)\in\Omega}\frac{\sum_{j=1}^{4}y^R_{mtj}}{4}.
\]

Requirement pass rates for each model are computed on the same set of 180 tasks. Because every task contains the same number of requirements, the task average equals the corresponding requirement-level micro-average; public and hidden requirements are aggregated separately. Capability analysis groups requirements by their primary capability assignments and computes micro-averages over the corresponding model–requirement records.

The all-record analysis comprises 1,980 generation records and 19,800 requirement slots. All 3,940 slots belonging to 393 gate-failed artifacts and one missing artifact receive zero, including the 10 slots for the missing artifact. The public-requirement pass rate is 3,947/11,880 = 33.2\%, and the hidden-requirement pass rate is 2,575/7,920 = 32.5\%. The gate-failure-or-missing-artifact rate is 394/1,980 = 19.9\%. The same algorithm is applied to all records within each grouping by source, difficulty, platform, or their intersections. An undefined denominator is recorded as missing rather than treated as a zero score.

\section{Human Cross-Check}

This section provides supplementary details on the human–machine requirement-level cross-check, including sample scope, label processing, and the distribution of indeterminate items.

\subsection{Sample and Implementation Records}

The human cross-check covers eight of the 75 Mobile player tasks and 67 artifacts. Matching is based on task, platform, model, artifact, and requirement text, and retains only items whose requirement wording is identical in the human sheet and the final evaluation list. After this alignment, 307 requirement judgments remain, covering 37 of the 80 (task, requirement) slots (1--9 per artifact, mean 4.6) and 10 of the 11 systems: one system contributes a single item and another none, and two tasks retain only one requirement each. The comparison is therefore a matched-item subsample, not a uniform ten-item sample.

The returned records provide the following information.

\par\medskip
\noindent\textbf{Information item}\par
Scope supported by available materials\par\medskip
\noindent\textbf{Task selection}\par
The agreement analysis covers eight Mobile player tasks. \par\medskip
\noindent\textbf{Artifact selection}\par
The returned table contains tasks, models, and anonymized identifiers, but does not provide itemized inclusion and exclusion rules for scored artifacts.\par\medskip
\noindent\textbf{Implementation records}\par
The standard specifies playtesting duration, immediate scoring, anchor calibration, and evidence retention. The returned table contains quality scores, descriptions, requirement judgments, and notes, but does not preserve playtesting duration per artifact, dual ratings, or complete recordings.\par\medskip

\subsection{Playtesting Protocol}

The quality protocol divides playtesting into cold start (0–1 minutes), core loop (1–5 minutes), and mid-to-late game (5–12 minutes), and specifies anonymous presentation, behavioral anchors, screen recording, and calibration. The current returned table contains nine quality dimensions; audio was not returned as a separate dimension. The human–machine agreement analysis in this paper uses only requirement-level labels. The scope of implementation records is described in Appendix F.1.

\subsection{Requirement Labels and Agreement-Rate Calculation}

Each artifact has only one human annotator. Humans label each requirement as "satisfied," "not satisfied," or "indeterminate." The quality-control field does not constitute a second independent human annotation. For the 241 judgeable human--machine label pairs in Table~\ref{tab:main-4}, Cohen's $\kappa$~\citep{cohen1960kappa} is approximately 0.26 (observed agreement: 149/241). The two requirement layers behave differently: $\kappa = 0.161$ for public requirements ($n = 151$; human 85.4\%, judge 47.0\%) and $\kappa = 0.430$ for hidden requirements ($n = 90$; human 60.0\%, judge 46.7\%). Disagreement is therefore concentrated in the public layer.

The 67 artifacts contain 307 aligned requirements, of which 66 are indeterminate (21.5\%), leaving 241 comparable records. The aligned requirements include 182 public requirements, of which 31 are indeterminate (17.0\%), and 125 hidden requirements, of which 35 are indeterminate (28.0\%). The denominator in Table~\ref{tab:main-4} is this judgeable set: the human–machine agreement rate is 149/241 = 61.8\%, the human satisfaction rate is 183/241 = 75.9\%, and the judge's pass rate is 113/241 = 46.9\%. Of these records, 81 are judged satisfied by humans but failed by the judge, and 11 are judged not satisfied by humans but passed by the judge. These rates are calculated only over aligned requirements that received human judgments and were judgeable.

\textbf{Rank preservation.} Before using these rates to bound the interpretation of the benchmark, we check whether the disagreement is uniform or order-dependent. The paired sample is uneven across systems (Appendix F.1): DeepSeek-V4-Pro contributes no paired item, and DeepSeek-V4-Flash and GLM-5.2 contribute fewer than 20 each. Table~\ref{tab:human-by-model} therefore compares the eight systems with at least 20 paired items. The judge is markedly stricter than human play-testing (46.9\% vs.\ 75.9\%, a gap of 29.0 percentage points), yet the two orderings agree almost perfectly (Spearman's $\rho = 0.95$). Per-system rates rest on 22--34 items and are therefore noisy; the aggregate matrix in Table~\ref{tab:main-4} remains the primary evidence. Absolute pass rates should accordingly be read as a strict lower bound on requirement satisfaction rather than as the share of features that work for a player, while differences between systems remain informative.

\begin{table}[H]
\caption{Per-system human and judge pass rates on the paired items of the human cross-check. Only systems with at least 20 paired items are listed; the judge column applies the all-record gate rule.}
\label{tab:human-by-model}\centering\small
\begin{tabularx}{\linewidth}{@{}>{\hsize=1.35\hsize}L >{\hsize=.85\hsize}L >{\hsize=.9\hsize}L >{\hsize=.9\hsize}L@{}}\toprule
System & Paired items & Human pass & Judge pass \\
\midrule
GPT-6-astra & 34 & 94.1\% & 61.8\% \\
Fable 5.1 & 33 & 84.8\% & 60.6\% \\
GLM-5.3 & 33 & 57.6\% & 33.3\% \\
Opus 5 & 31 & 90.3\% & 58.1\% \\
Fable 5 & 28 & 42.9\% & 32.1\% \\
GPT-5.5 & 25 & 80.0\% & 48.0\% \\
GPT-5.6-sol & 24 & 83.3\% & 50.0\% \\
DeepSeek-V4.1-Flash & 22 & 81.8\% & 40.9\% \\
\bottomrule\end{tabularx}\end{table}

\section{Statistical and Sensitivity Analyses}

\subsection{Requirement-Difficulty Stratification}

Each difficulty stratum contains 450 requirements and 11 models, for a total of 4,950 model–requirement records.

\begin{table}[H]
\caption{Requirement pass rates across difficulty strata.}
\label{tab:difficulty}\centering\small
\begin{tabularx}{\linewidth}{*{6}{L}}\toprule
Difficulty & Requirements & Model–requirement records & Public pass rate & Hidden pass rate & Overall pass rate \\
\midrule
Easy & 450 & 4,950 & 38.75\% & 40.43\% & 39.54\% \\
Medium & 450 & 4,950 & 32.69\% & 31.64\% & 32.22\% \\
Hard & 450 & 4,950 & 31.80\% & 29.18\% & 30.75\% \\
Extreme & 450 & 4,950 & 30.70\% & 25.65\% & 29.25\% \\
\bottomrule\end{tabularx}\end{table}

Spearman's $\rho$ between $a_{DS}$ and the 11-model average pass rate of each requirement is -0.1201; when computed over the 19,800 individual model–requirement records, Spearman's $\rho$ is -0.0854.

\begin{table}[H]
\caption{Overall requirement pass rates within each subset across the global requirement-difficulty strata.}
\label{tab:difficulty-subsets}\centering\small
\begin{tabularx}{\linewidth}{*{5}{L}}\toprule
Subset & Easy & Medium & Hard & Extreme \\
\midrule
Player PC & 18.34\% & 20.01\% & 20.02\% & 20.48\% \\
Player Mobile & 37.80\% & 34.67\% & 35.95\% & 36.69\% \\
Reconstruction & 56.91\% & 49.12\% & 47.92\% & 49.73\% \\
Expert & 72.11\% & 63.30\% & 65.66\% & 72.73\% \\
\bottomrule\end{tabularx}\end{table}

The correlations between $a_{DS}$ and requirement-level 11-model average pass rates within player PC, player Mobile, reconstruction, and expert subsets are -0.0023, -0.0216, -0.0800, and -0.0118, respectively. Weighting the four subset-specific Extreme-minus-Easy differences by their fixed shares of the 1,800 requirements gives -0.33 percentage points. The reduction from a pooled Extreme-minus-Easy difference of -10.28 percentage points to -0.33 after fixing subset proportions identifies subset composition as an important contributor to the aggregate gradient. The aggregate gradient thus combines differences in subset composition with weak associations within subsets. Joint reporting by difficulty, source, and track makes these two aspects of benchmark performance visible.

This study conducts descriptive analysis of a fixed sample and does not report significance tests or confidence intervals. Only the final record is retained for each model–task combination. Requirement records share tasks, models, and the evaluator; the 19,800 requirement slots are therefore not treated as independent and identically distributed observations.

Stratified results reported in the paper cover model, source, capability, difficulty, platform, and common task set, with separate numerators and denominators for each grouping. Appendices G.2--G.4 provide capability-assignment and aggregation sensitivity analyses, shared-source sensitivity analysis across platforms, and common gate-passing task sets, respectively.

\subsection{Capability Assignment and Aggregation Sensitivity}

The capability-sensitivity analysis uses two alternative protocols. Under the multilabel protocol, a requirement is counted under all its available capability labels, so capability denominators are no longer mutually exclusive. Under the no-self-evaluation protocol, Astra is removed before recalculation. Each model has the same requirement denominator for a given capability; consequently, averaging capability pass rates first within each model is equivalent to the requirement-level micro-average.

Of the 1,800 requirements, 803 have one capability label and 997 have two.  Capability groups vary in coverage. C7 is the smallest primary-capability group, with 85 requirements and 935 requirement records across the 11 models.

The alternative protocols produce the same broad capability pattern as the main analysis. Under multilabel aggregation, pass rates range from 26.6\% for C2 to 34.6\% for C1, with C7 at 28.1\% and C3 at 33.8\%. After excluding the same-model self-evaluation condition, primary-capability pass rates range from 25.16\% for C2 to 35.8\% for C1, with C7 nearly tied at 25.18\% and C3 at 35.4\%. Thus, C1 and C3 remain the highest capabilities and C2 and C7 remain the lowest under both sensitivity checks, although the exact ordering of the middle capabilities changes. The C7 estimate should be interpreted cautiously because it rests on only 85 requirements (935 records across the 11 models; 850 after excluding Astra).

\subsection{Shared-Source Sensitivity Analysis across Platforms}

PC includes 10 platform-specific expert-designed tasks, whereas Mobile includes none. To examine whether this source difference determines the overall platform results, we exclude expert-designed tasks and retain only player tasks and commercial-mechanics reconstruction tasks, which are shared source types across the two platforms. After this exclusion, PC and Mobile each contain 75 player tasks and 10 commercial-mechanics reconstruction tasks and thus have identical source compositions; no further source reweighting is necessary.

On the shared-source support set of 935 model–task records per platform, the gate-failure-or-missing-artifact rates for PC and Mobile are 29.41\% and 12.41\%, Completeness is 23.94\% and 38.63\%, and Richness is 18.42\% and 41.76\%, respectively. The direction of higher values for Mobile persists after excluding the source unique to PC. This analysis controls only the numerical composition of sources; the two platforms still differ in their specific Game Specifications, interaction paths, requirement content, and other task attributes.

\subsection{Common Task Set}

All-record requirement pass rates are calculated over the same universe of 180 final tasks for every model. This common task universe should be distinguished from the gate-passing support used for conditional comparisons. Intersecting the included task sets across all 11 models leaves 103 tasks, corresponding to 1,133 model--task records; on this intersection, the public pass rate is 43.2\% (2,934/6,798) and the hidden pass rate is 42.2\% (1,914/4,532). For pairwise comparisons, the intersection is recomputed separately for each model pair: the 55 pairwise supports contain between 121 and 152 tasks, with a median of 137 tasks. Therefore, the pairwise comparisons use common support within each pair, but the support set is not identical across all 55 pairs. Intersections are jointly determined by artifact existence and gate outcomes.

\section{Supplementary Materials Index}

The unified public entry point for OCTAVE is the \href{https://github.com/rezona-ai/octave-bench}{GitHub repository}. Repository links below refer to files or directories on the \texttt{main} branch.

\par\medskip
\noindent\textbf{Benchmark task files}\par
The repository provides all 180 benchmark tasks in \href{https://github.com/rezona-ai/octave-bench/blob/main/data/octave\_180\_en.jsonl}{octave\_180\_en.jsonl} (English) and \href{https://github.com/rezona-ai/octave-bench/blob/main/data/octave\_180\_zh.jsonl}{octave\_180\_zh.jsonl} (Chinese), including 150 player tasks, 20 commercial-mechanics reconstruction tasks, and 10 expert-designed tasks.\par\medskip
\noindent\textbf{Dataset documentation}\par
The \href{https://github.com/rezona-ai/octave-bench/blob/main/data/README.md}{GitHub dataset card} documents the released files, task fields, and source composition for the reported experiments.\par\medskip
\noindent\textbf{Complete static-evaluation templates}\par
The four PC/Mobile and 2D/3D templates under \href{https://github.com/rezona-ai/octave-bench/tree/main/llm-eval}{llm-eval}.\par\medskip
\noindent\textbf{Generation protocol and configurations}\par
\href{https://github.com/rezona-ai/octave-bench/blob/main/harness/run\_bench.py}{run\_bench.py}, \href{https://github.com/rezona-ai/octave-bench/blob/main/configs/models.json}{model configurations}, the \href{https://github.com/rezona-ai/octave-bench/tree/main/configs}{generation-prompt template directory}, and the \href{https://github.com/rezona-ai/octave-bench/blob/main/docs/reproduce.md}{run instructions}.\par\medskip
\noindent\textbf{Automated evaluation and aggregation implementation}\par
\href{https://github.com/rezona-ai/octave-bench/blob/main/eval/run\_eval.py}{run\_eval.py}, \href{https://github.com/rezona-ai/octave-bench/blob/main/utils/static\_eval.py}{static\_eval.py}, and \href{https://github.com/rezona-ai/octave-bench/blob/main/scoring/aggregate.py}{aggregate.py}. These engineering entry points do not replace the paper-specific statistical scripts.\par


\medskip

\begin{thebibliography}{99}
\bibitem[Zubek(2020)]{zubek2020elements} Zubek, R. \emph{Elements of Game Design}. MIT Press, 2020.
\bibitem[Swink(2008)]{swink2008gamefeel} Swink, S. \emph{Game Feel: A Game Designer's Guide to Virtual Sensation}. Morgan Kaufmann, 2008.
\bibitem[Chen et al.(2021)]{chen2021codex} Chen, M. et al. Evaluating Large Language Models Trained on Code. arXiv:2107.03374, 2021.
\bibitem[Austin et al.(2021)]{austin2021program} Austin, J. et al. Program Synthesis with Large Language Models. arXiv:2108.07732, 2021.
\bibitem[Liu et al.(2024)]{liu2023repobench} Liu, T. et al. RepoBench: Benchmarking Repository-Level Code Auto-Completion Systems. ICLR, 2024.
\bibitem[Jimenez et al.(2024)]{jimenez2024swebench} Jimenez, C. E. et al. SWE-bench: Can Language Models Resolve Real-World GitHub Issues? ICLR, 2024. \url{https://openreview.net/forum?id=VTF8yNQM66}.
\bibitem[Zhao et al.(2025)]{zhao2024commit0} Zhao, W. et al. Commit0: Library Generation from Scratch. ICLR, 2025.
\bibitem[Starace et al.(2025)]{starace2025paperbench} Starace, G. et al. PaperBench: Evaluating AI's Ability to Replicate AI Research. ICML, 2025.
\bibitem[Ding et al.(2026)]{nl2repo2026} Ding, J. et al. NL2Repo-Bench: Towards Long-Horizon Repository Generation Evaluation of Coding Agents. arXiv:2512.12730, 2026.
\bibitem[Hendrikx et al.(2013)]{hendrikx2013pcg} Hendrikx, M. et al. Procedural Content Generation for Games: A Survey. ACM TOMM, 2013.
\bibitem[Summerville et al.(2018)]{summerville2018pcgml} Summerville, A. et al. Procedural Content Generation via Machine Learning (PCGML). IEEE Transactions on Games, 2018.
\bibitem[Chen et al.(2023)]{chen2023gamegpt} Chen, D. et al. GameGPT: Multi-agent Collaborative Framework for Game Development. arXiv:2310.08067, 2023.
\bibitem[Wu et al.(2024)]{wu2024idge} Wu, H. et al. Instruction-Driven Game Engines on Large Language Models. arXiv:2404.00276, 2024.
\bibitem[Zhang et al.(2026b)]{rpgagent2026} Zhang, S. et al. RPGAgent: Driving Coherent Story-to-Play Generation with an LLM-Based Multi-Agent System. CHI, 2026.
\bibitem[Jiang et al.(2026)]{opengame2026} Jiang, Y. et al. OpenGame: Open Agentic Coding for Games. arXiv:2604.18394, 2026.
\bibitem[Chi et al.(2026)]{gamedevbench2026} Chi, W. et al. GameDevBench: Evaluating Agentic Capabilities Through Game Development. arXiv:2602.11103, 2026.
\bibitem[Zhang et al.(2025)]{vgamegym2025} Zhang, W. et al. V-GameGym: Visual Game Generation for Code Large Language Models. arXiv:2509.20136, 2025.
\bibitem[Jia et al.(2026)]{gamegenverifier2026} Jia, C. et al. GameGen-Verifier: Parallel Keypoint-Based Verification for LLM-Generated Games via Runtime State Injection. arXiv:2605.07442, 2026.
\bibitem[Luo et al.(2026)]{gamecraftbench2026} Luo, T. et al. GameCraft-Bench: Can Agents Build Playable Games End-to-End in a Real Game Engine? arXiv:2606.17861, 2026.
\bibitem[Zhang et al.(2026c)]{webgamebench2026} Zhang, W. et al. WebGameBench: Requirement-to-Application Evaluation for Coding Agents via Browser-Native Games. arXiv:2605.17637, 2026.
\bibitem[Chen et al.(2026)]{gamexpertbench2026} Chen, K. et al. GameXpert-Bench: How Far Are Coding Agents from Expert Game Development? arXiv:2608.21833, 2026.
\bibitem[La et al.(2026)]{gameenginebench2026} La, B. et al. GameEngineBench: Evaluating Coding Agents on Real C++ Runtime Environments. arXiv:2607.03525, 2026.
\bibitem[Zhang et al.(2026a)]{gameasgbench2026} Zhang, X. et al. GameASG-Bench: Benchmarking Autonomous Software Generation for Game Development. arXiv:2609.21293, 2026.
\bibitem[Che et al.(2026)]{gamelogicbench2026} Che, X. et al. GameLogicBench: Evaluating Coding Agents on Runtime Game Logic with Tick-Level State Assertions. arXiv:2609.21562, 2026.
\bibitem[Jain et al.(1984)]{jain1984fairness} Jain, R. K., Chiu, D.-M. W., and Hawe, W. R. A Quantitative Measure of Fairness and Discrimination for Resource Allocation in Shared Computer Systems. Digital Equipment Corporation, Technical Report TR-301, 1984. \url{https://arxiv.org/abs/cs/9809099}.
\bibitem[Spearman(1904)]{spearman1904} Spearman, C. The Proof and Measurement of Association between Two Things. \emph{The American Journal of Psychology}, 15(1):72--101, 1904.
\bibitem[Cohen(1960)]{cohen1960kappa} Cohen, J. A Coefficient of Agreement for Nominal Scales. \emph{Educational and Psychological Measurement}, 20(1):37--46, 1960.
\bibitem[Yang et al.(2024)]{yang2024sweagent} Yang, J. et al. SWE-agent: Agent-Computer Interfaces Enable Automated Software Engineering. arXiv:2405.15793, 2024. \url{https://arxiv.org/abs/2405.15793}.
\bibitem[Wang et al.(2024)]{wang2024openhands} Wang, X. et al. OpenHands: An Open Platform for AI Software Developers as Generalist Agents. arXiv:2407.16741, 2024. \url{https://arxiv.org/abs/2407.16741}.
\bibitem[Zheng et al.(2023)]{zheng2023llmjudge} Zheng, L. et al. Judging LLM-as-a-Judge with MT-Bench and Chatbot Arena. NeurIPS, 2023. \url{https://arxiv.org/abs/2306.05685}.
\bibitem[Panickssery et al.(2024)]{panickssery2024selfpref} Panickssery, A., Bowman, S. R., and Feng, S. LLM Evaluators Recognize and Favor Their Own Generations. NeurIPS, 2024. \url{https://arxiv.org/abs/2404.13076}.
\bibitem[DeepSeek-AI(2026a)]{deepseekv42026} DeepSeek-AI. DeepSeek-V4: Towards Highly Efficient Million-Token Context Intelligence. arXiv:2606.19348, 2026. \url{https://arxiv.org/abs/2606.19348}.
\bibitem[DeepSeek-AI(2026b)]{deepseekv41flash2026} DeepSeek-AI. DeepSeek-V4.1-Flash: Pushing the Limits of KV Cache Compression. arXiv:2609.19969, 2026. \url{https://arxiv.org/abs/2609.19969}.
\bibitem[Anthropic(2026a)]{fable52026} Anthropic. System Card: Claude Fable 5 and Claude Mythos 5. June 2026. \url{https://www-cdn.anthropic.com/57a52ea7d8f0e54e8a542e908266086df425cdf5/Claude%20Fable%205%20&%20Claude%20Mythos%205%20System%20Card.pdf}.
\bibitem[Anthropic(2026b)]{fable512026} Anthropic. System Card: Claude Fable 5.1 and Claude Mythos 5.1. September 2026. \url{https://www-cdn.anthropic.com/0339e6a7c5c7b87f5c07798616dc32c215d14235/Claude%20Fable%205.1%20%26%20Claude%20Mythos%205.1%20System%20Card.pdf}.
\bibitem[Anthropic(2026c)]{opus52026} Anthropic. System Card: Claude Opus 5. July 2026. \url{https://www-cdn.anthropic.com/ceaf5c7ff2783855203fde8208ec311252dced5b/Claude%20Opus%205%20System%20Card.pdf}.
\bibitem[Z.ai(2026a)]{glm522026} Z.ai. GLM-5.2: Built for Long-Horizon Tasks. Z.ai Blog, June 2026. \url{https://z.ai/blog/glm-5.2}.
\bibitem[Z.ai(2026b)]{glm532026} Z.ai. GLM-5.3: Frontier Coding with Emergent Cyber Capabilities. Z.ai Blog, August 2026. \url{https://z.ai/blog/glm-5.3}.
\bibitem[OpenAI(2026a)]{gpt552026} OpenAI. GPT-5.5 System Card. April 2026. \url{https://deploymentsafety.openai.com/gpt-5-5/gpt-5-5.pdf}.
\bibitem[OpenAI(2026b)]{gpt562026} OpenAI. GPT-5.6 System Card. July 2026. \url{https://deploymentsafety.openai.com/gpt-5-6/gpt-5-6.pdf}.
\bibitem[OpenAI(2026c)]{gpt6astra2026} OpenAI. GPT-6 Astra System Card. September 2026. \url{https://deploymentsafety.openai.com/gpt-6-astra/gpt-6-astra.pdf}.
\bibitem[Anthropic(2025)]{claudecode2025} Anthropic. Claude 3.7 Sonnet and Claude Code. Anthropic News, February 2025. \url{https://www.anthropic.com/news/claude-3-7-sonnet}.
\bibitem[OpenAI(2025)]{codexcli2025} OpenAI. Introducing Codex. May 2025. \url{https://openai.com/index/introducing-codex/}.
\bibitem[Chen(2026)]{codexharness2026} Chen, C. Unlocking the Codex Harness: How We Built the App Server. OpenAI, February 2026. \url{https://openai.com/index/unlocking-the-codex-harness/}.
\end{thebibliography}
\end{document}